\documentclass{article}

\usepackage{PRIMEarxiv}

\usepackage{comment}
\usepackage[utf8]{inputenc} 
\usepackage[T1]{fontenc}    
\usepackage{hyperref}       
\usepackage{url}            
\usepackage{booktabs}       
\usepackage{amsfonts}       
\usepackage{nicefrac}       
\usepackage{microtype}      
\usepackage{lipsum}
\usepackage{fancyhdr}       
\usepackage{graphicx}       
\usepackage{subcaption}
\usepackage[inkscapeformat=png]{svg}
\graphicspath{{media/}}     
\usepackage{multirow}
\usepackage{pifont}
\usepackage{xcolor}

\title{Self-supervised vision-langage alignment of deep learning representations for bone X-rays analysis
}

\author{Alexandre Englebert$^{1,2}$, Anne-Sophie Collin$^{1}$, Olivier Cornu$^{2}$ \& Christophe De Vleeschouwer$^{1}$\\~\\
$^{1}$Information and Communication Technologies, Electronics and Applied Mathematics (ICTEAM),\\ UCLouvain, Belgium\\
$^{2}$Service de chirurgie orthopédique et traumatologie, cliniques universitaires Saint-Luc UCL,\\
Neuro Musculo Skeletal Lab (NMSK), Institut de Recherche Expérimentale et Clinique (IREC),\\
UCLouvain, Brussels, Belgium
}

\begin{document}
\maketitle

\begin{abstract}

This paper proposes leveraging vision-language pretraining on bone X-rays paired with French reports to address downstream tasks of interest on bone radiography. A practical processing pipeline is introduced to anonymize and process French medical reports. Pretraining then consists in the self-supervised alignment of visual and textual embedding spaces derived from deep model encoders. The resulting image encoder is then used to handle various downstream tasks, including quantification of osteoarthritis, estimation of bone age on pediatric wrists, bone fracture and anomaly detection. Our approach demonstrates competitive performance on downstream tasks, compared to alternatives requiring a significantly larger amount of human expert annotations. Our work stands as the first study to integrate French reports to shape the embedding space devoted to bone X-Rays representations, capitalizing on the large quantity of paired images and reports data available in an hospital. By relying on generic vision-laguage deep models in a language-specific scenario, it contributes to the deployement of vision models for wider healthcare applications.

\end{abstract}


\section{Introduction}

In the medical domain, particularly in radiography, large-scale datasets are generally limited to English reports and to specific body areas. To the best of our knowledge, the only large publicly available radiography-report dataset is MIMIC-CXR\cite{johnson2019mimic}, containing 377,110 Chest Xray images and their corresponding free-text reports in English. This raises a significant challenge when applying the models derived from those data to images other than Chest Xrays.

Moreover, privacy regulations such as the General Data Protection Regulation (GDPR)\cite{GDPR} impose strict limitations on the distribution and sharing of medical databases containing sensitive patient information.

To address this limitation, one viable approach would be to utilize local data available within a given hospital or healthcare institution. Hospitals typically maintain their own databases of medical images and associated reports, which are collected as part of routine clinical practice. While these local datasets may not be as extensive as publicly available datasets, they still contain valuable information that can be leveraged for training and evaluating machine learning models.

Therefore, in this paper, we propose to explore \textbf{vision-language pretraining} using \textbf{bone X-rays} paired with \textbf{French reports} sourced from a \textbf{single university hospital} department. Specifically, our work aims at aligning deep embedding representations of Bone X-Rays and French Reports for solving image-based medical tasks with limited
annotation.


The main contributions are:

\begin{itemize}
    \item We leveraged bone radiographs and their associated French reports from a single hospital to pretrain a versatile vision-language model, to be used as a backbone for a variety of tasks trained with limited supervision. The obtained multi-modal representation is shown to result in downstream task performance that are competitive with models trained with a significantly larger amount of human supervision.
    \item We demonstrate the feasibility of vision-language pretraining in the medical field, using other language than English.
    \item We present a practical guide on harnessing medical data from an individual hospital, detailing our approach, with a special emphasis on the anonymization process. In particular, a French adaptation of the DEDUCE~\cite{menger2018deduce} method was developped and is made available\footnote{\url{https://github.com/aenglebert/deduced}} to facilitate the pseudonymization of medical reports in French.
\end{itemize}

\section{Related Work}\label{sec:related_work}

\subsection{Self-Supervised Learning}

The role of self-supervision is to learn effective reprensations by capturing inherent pattern in data rather than relying on human-based annotations. Various techniques are available and tailored to specific modalities. In the context of this study, the pertinent modalities are imagery and text, which will be the primary focus.

In the case of texts, two self-supervised methodologies stand out: masked text modeling (like BERT~\cite{devlin2018bert} or RoBERTa\cite{liu2019roberta}), where a bidirectional transformer is trained to predict a masked word within a sentence based on the contextual cues, and generative pretraining (with GPT-like models~\cite{radford2018improving, brown2020language}), which predicts the next word in a given sentence.

In computer vision, a prevalent approach involves encouraging different versions of the same image to have similar network embeddings, either by using a contrastive loss (e.g. SimCLR~\cite{simclr}, MOCO~\cite{moco}), a self-distillation siamese network (e.g. DINO~\cite{caron2021emerging}, BYOL~\cite{grill2020bootstrap}), or a canonical correlation based method (such as Barlow Twins~\cite{zbontar2021barlow} or SWAV~\cite{caron2020swav}). Typically, these techniques generate diverse versions of an image via data augmentation. Inspired from masked language modeling, masked image modeling is another kind of self-supervision that works by predicting masked patches of an image (e.g. Masked Autoencoder~\cite{he2022masked}, SimMIM~\cite{xie2022simmim}).

In our work, both text and image modalities are considered. Vision-Language Pretraining (VLP) is a particular type of self-supervised learning where the image and text representations are trained together, leveraging the relation between a given image and a related text. 
This is typically done using contrastive learning (CLIP~\cite{radford2021learning}, ALIGN~\cite{jia2021scaling}, DeCLIP~\cite{li2021supervision}, GLIP~\cite{li2022grounded}), pseudo-tasks such as alignment prediction and masked multi-modal modeling (VilBERT~\cite{lu2019vilbert}, VisualBERT~\cite{li2019visualbert}), image captioning (SimVLM~\cite{wang2021simvlm}, Virtex~\cite{desai2021virtex}), or a combination of multiple uni-modal and multi-modal self-supervisions (Flava~\cite{singh2022flava}). The fusion between the modalities can be performed at an early stage, with a common encoder capable of handling both text and images (like VilBERT~\cite{lu2019vilbert} or VisualBERT~\cite{li2019visualbert}), or at a late stage with two separate encoders like CLIP~\cite{radford2021learning}, or any kind of intermediate~\cite{li2021align, singh2022flava}. The training is usually performed on large scale datasets of images with their respective captions or description (from a few millions to 1.8B for ALIGN~\cite{li2021align}).

\subsection{Medical applications of self-supervised Vision-Language Pretraining}\label{sec:med_vlp}

Multiple adaptations of VLP methods to clinical datasets have been envisioned.
ConVIRT~\cite{Zhang2022convirt} is a precursor in the application of contrastive VLP.
Recent advancements have extended beyond global contrastive alignment between image and text by the incorporation of local alignment as exemplified by GLORIA~\cite{huang2021gloria}, LoVT~\cite{muller2022joint}, MGCA~\cite{wang2022multi} or PRIOR~\cite{cheng2023prior}. Since our approach primarily questions the benefit of running VLP on French reports, solely employs global contrastive loss and training based on ConVIRT.

Most previous works on medical VLP have been validated on chest radiographs, using (Bio)ClinicalBERT~\cite{alsentzer2019publicly} as a text encoder. (Bio)ClinicalBERT has been trained on medical reports from the MIMIC III dataset~\cite{johnson2016mimic} and shows superior performance in comparison to a biomedical model like BioBERT~\cite{lee2020biobert}, trained on biomedical domain corpora such as PubMed abstract and PMC full-text articles. Both ClinicalBERT and BioBERT are trained on English texts.

In contrast, our work considers French documents and bone radiography. This poses multiple challenges since, at the time of writing, many useful tools are English-only (e.g. CheXpert labeler~\cite{irvin2019chexpert}, RadGraph~\cite{jain2021radgraph}, negBIO~\cite{peng2018negbio}). 

To exploit French medical documents in the frame of a vision-language pretraining, our work considers two alternatives, corresponding to French-only models and multi-lingual models. 

On the one hand, French-only models, such as CamemBERT~\cite{martin2019camembert}, have been adapted to the biomedical domain with models like Dr BERT~\cite{labrak2023drbert} or CamemBERT-BIO~\cite{touchent2023camembertbio}. However, the amount of data used for training these models is smaller than their English counterparts (e.g., 4.5B + 13.5B words for BioBERT, 3.1B words for PubMedBERT~\cite{gu2021domain}, versus 1B for NACHOS used by DrBERT, and 413M for CamemBERT-bio).

On the other hand, multimodal languages such as mBERT~\cite{devlin2018bert}, XLM-Roberta (XLMR)\cite{conneau2019xlm}, or MLUKE\cite{ri2021mluke} benefit from having more data for pretraining than French-only models. Moreover, they enable cross-lingual transfer of knowledge. Methods like Self-alignment pretraining (Sap)~\cite{liu2021learning} have been applied to multilingual general models such as XLMR~\cite{conneau2019xlm}, showing promising performance on tasks such as Biomedical Entity Linking. This pretraining involves aligning the embeddings of synonyms of concepts from the Unified Medical Language System (UMLS), a compendium that integrates and harmonizes various medical terminologies and classifications. Models of this kind become particularly valuable when biomedical resources for a specific language are scarce.

\section{Methodology}

\begin{figure}
    \centering

    \includegraphics[width=\linewidth]{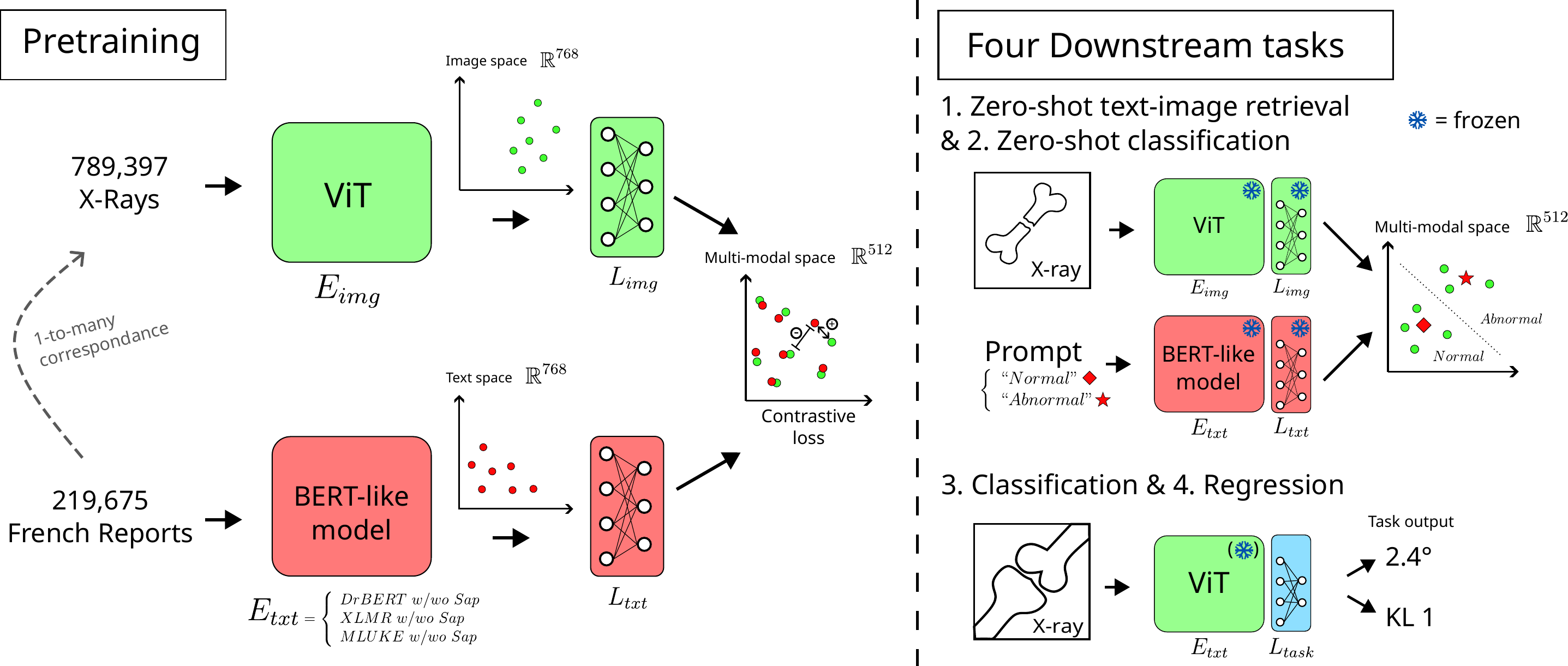}
    \caption{\textbf{General overview:} Vision-Language Pretraining (VLP) consists in the alignment of the embeddings for both X-Rays and French Reports. Once pretrained, the encoders can be adapted to different downstream tasks.}
    \label{fig:vlp}
\end{figure}

This section presents the methodology that starts with the Vision-Language Pretraining in Section~\ref{sec:vlp_pretraining}, followed by the evaluation in different downstream tasks described in Section~\ref{sec:downstream_tasks}. A visual representation is presented in Figure~\ref{fig:vlp}.

\subsection{Learning joint representations for clinical reports and X-ray images}\label{sec:vlp_pretraining}

This section describes how the representation of medical images can be adapted to fit the representation adopted for (French) clinical reports. A general overview of our setup is provided in Section~\ref{sec:vlp_setup}. Section~\ref{sec:text_encoder} describes the various text encoders considered in our study. While Section~\ref{sec:vlp_resolution} investigates different techniques to increase the resolution of an off-the-shelf visual encoder to get closer to the one of our images.

\subsubsection{Vision-language pretraining (VLP)}\label{sec:vlp_setup}

In this work, we employed a traditional bi-encoder global contrastive framework, analogous to that proposed by ConVIRT~\cite{Zhang2022convirt}. Previous works such as ConVIRT~\cite{Zhang2022convirt}, GLORIA~\cite{huang2021gloria}, MGCA~\cite{wang2022multi} or PRIOR~\cite{cheng2023prior} utilized a (Bio)ClinicalBERT~\cite{alsentzer2019publicly} as the text encoder and ResNet50~\cite{he2016deep} pretrained on ImageNet~\cite{ILSVRC15} as initialization for the image encoder. However, this text encoder is designed for English language and is consequently not ideally suited to the reports of our hospital. Therefore, we explored French and multilingual alternatives to define our text encoder. For the image encoder component, we opted for the more recent ViT~\cite{dosovitskiy2020image} model instead of a ResNet.
More details about the explored text and image encoders are provided in Section~\ref{sec:text_encoder} and Section~\ref{sec:vlp_resolution}, respectively.
The output CLS (classification token) of the image and text encoders serves as a global representation of the image and text, respectively, and are each linearly projected as a 512-dimensional vector. The objective of the Vision-Language Pretraining is to bring closer the representation of images to the representation of the corresponding report by fine-tuning both image and text encoders. In practice, a CLIP loss, as described by Radford et al~\cite{radford2021learning}, is employed to minimize the cosine distance between image and text vectors from the same study, while simultaneously reducing the distance between text and image vectors from different studies. In our work, a study denotes the outcomes of a radiological examination. Hence, it is specific to one patient and to one visit to the hospital, and includes one report and potentially multiple X-ray images.


In preliminary experiments, we evaluated the effectiveness of using one random image from each study compared to utilizing all images from the study and pooling the results (either by averaging the CLS tokens or using attention pooling). It revealed that, the benefits of using multiple images pretraining were not clearly evident, while the complexity of the framework increased. Consequently, we opted to randomly select one image for each study in a batch.

The source code for the vision-language pretraining is available at \url{https://github.com/aenglebert/multimodal_bone}.

\subsubsection{Text encoder and self-alignment pretraining (Sap)}\label{sec:text_encoder}

We selected three candidate text encoders:
\begin{itemize}
    \item XLM-Roberta (XLMR)~\cite{conneau2019xlm}, which is a multi-lingual text encoder based on Roberta~\cite{liu2019roberta} and trained using Masked Language Modeling using texts in 100 languages.
    \item MLUKE~\cite{ri2021mluke}, a multi-lingual version of LUKE~\cite{yamada2020luke} trained with Masked Language Modeling and Masked Entity Prediction on 24 languages.
    \item Dr BERT~\cite{labrak2023drbert}, a french encoder based on CamemBERT~\cite{martin2019camembert} and trained using Masked Language Modeling on a French biomedical corpus.
\end{itemize}

We also augmented the comparison with the self-alignment pretraining method (Sap)~\cite{liu2021learning}. In this method, a pretraining consists in the alignment of the embeddings of synonymes of concepts from the Unified Medical Language System (UMLS), a compendium that integrates and harmonizes diverse medical terminology and classifications. For XLMR, the original XLMR SapBERT model was used, and we pretrained MLUKE and Dr BERT using the source code of the authors with the parameters described in their paper and UMLS 2020AA, as implemented in the original code.

\subsubsection{Image Encoder}\label{sec:vlp_resolution}

The image encoder was initialised from a ViT B16 224x224 pretrained on ImageNet~\cite{dosovitskiy2020image}.
Multiple resolutions have been explored, 224x224 as the native resolution of the model, and resolutions increased to 336x336 and 448x448.
To increase the resolution of the image encoder, the 224x224 image encoder is first pretrained with Vision-Language pretraining as explained in Section~\ref{sec:vlp_pretraining} and then modified in two possible ways:
\begin{itemize}
    \item Interpolation of the position embeddings~\cite{dosovitskiy2020image} (named ViT B16 336 and ViT B16 448 in the following). This technique involves the interpolation of the trained position embeddings to enable the input of a greater number of tokens to the vision transformer, thereby accommodating images with higher resolutions, all while preserving their semantic significance.
    \item Increasing of the patch size using the pseudoinverse resizing methods described in FlexiVit~\cite{beyer2023flexivit}. This approach ensures the retention of the original number of tokens, while each token covers a larger number of input pixels. Unlike FlexiVit, the resizing is performed once to initialize a ViT B24 336 and a ViT B32 448 models. This method requires less computation in comparison to interpolating the position embeddings.
\end{itemize}

The vision-language pretraining is then continued with images of increased resolution.



\subsection{Downstream tasks}\label{sec:downstream_tasks}

This section introduces the different downstream tasks considered to evaluate the performance of the pretraining described in Section~\ref{sec:vlp_pretraining}.

The first evaluation described in Section~\ref{sec:validation} is performed on a validation dataset derived from the VLP dataset.

More extensive downstream on over datasets and tasks encompass both trained tasks (Section~\ref{sec:trained_tasks}) and zero-shot tasks (Section~\ref{sec:zero_tasks}). The source code for the different downstream tasks is available at \url{https://github.com/aenglebert/ortho_vlp_eval}.

\subsection{Tasks trained on images captured in the same hospital than the VLP ones}\label{sec:validation}
In order to quantify the supervised performances of the pretrained model on images from the same hospital as the images used during the VLP process, we have created a small supervised dataset for fracture detection. The labels of this dataset were automatically produced by processing reports unseen during VLP training using a Llama3 70B model~\cite{touvron2023llama}. The details of the construction of this dataset are provided in Appendix~\ref{app:val_dataset}.

A single linear classification layer is trained on top of each frozen encoder. The training is performed using random subsets of the training data of various sizes, 8 times per model for a given number of training images to allow the computation of a confidence interval.

\subsubsection{Tasks trained on images captured outside the hospital considered by VLP}\label{sec:trained_tasks}

For trained task, we adopted an evaluation strategy similar to previous studies~\cite{huang2021gloria, muller2022joint, wang2022multi, cheng2023prior}, with two settings: linear classification on a frozen image encoder, and full fine-tuning. Two tasks are performed depending on the datasets, classification or regression.
To assess data efficiency, we compared training using either the entire training set or a smaller part of the training set (from 1 to 10\%).

The objective was to evaluate performance in relation to pretraining and not to obtain maximal performance on the downstream tasks per se. 
For the linear evaluation, a single linear layer was appended to the CLS token of the image encoder to facilitate classification.

\subsubsection{Zero-Shot Tasks}\label{sec:zero_tasks}

For the zero-shot tasks, the vision-language pretrained models are utilized without additional fine-tuning. Two tasks are investigated: zero-shot classification and zero-shot retrieval. 
In zero-shot classification, a text prompt is classically associated to each class, and images are assigned to the class whose text prompt embedding is the closest (in cosine distance) to the image embedding. 
In zero-shot retrieval, a fixed number of images with the closest embedding from a class text prompt embedding are retrieved.
In practice, this is achieved by leveraging a measurement of distance between the projected CLS (classification token) from both encoders in the multi-modal space, specifically, for the image under consideration and for the reference prompt associated with a given class. 


Four prompting strategies were kept for the evaluation:

\begin{itemize}
    \item \textbf{Text binary}: A simple prompt with the name of the target class is used. The negative prompt being a "normal" prompt.
    \item \textbf{Text enumeration}: The class prompt is constructed as a comma separated list of sub-classes of the target class.
    \item \textbf{Latent minimum}: The same sub-classes are encoded as separate prompts by the text encoder. Multiple embedding thus exists that belongs to the same target class.
    \item \textbf{Latent mean}: The same sub-classes are encoded separately by the text encoder. The target class embedding is produced by averaging the sub-classes embedding.
\end{itemize}

For the classification task, the predicted class is assigned based on the distance in the multi-modal space between the image and a reference prompt (cfr. zero-shot downstream tasks in Figure~\ref{fig:vlp}).


For the retrieval task, the top k images with the lowest distance from the text query in the multi-modal space are retrieved. Precision is computed across various values of k. This evaluation is closer to zero-shot experiments presented in previous works on the CheXpert 8x200 dataset~\cite{Zhang2022convirt} for Chest X-Rays. 


\section{Experimental validation}\label{sec:experiments}

\subsection{Vision-language pretraining on Bone X-Rays and French Reports}\label{sec:exp:vlp}

The Vision-Language pretraining (VLP) described in Section~\ref{sec:vlp_pretraining} was performed on a dataset of 219,675 studies (each study corresponding to one radiological examination) composed of one French Report per study coupled to one or multiple X-rays, with 789,397 individual X-ray images in total. 
This extensive dataset was curated from 75600 patients with follow-ups at a single university hospital's orthopedic surgery department over a span of 20 years. Its creation involved significant steps, notably pseudonymization using a modified version of the DEDUCE~\cite{menger2018deduce} framework, originally developed in Dutch and adapted to French for the needs of our work (the source code is available at \url{https://github.com/aenglebert/deduced} ). To the best of our knowledge, this dataset, featuring bone X-ray images from one hospital alongside French reports, has no equivalent in current literature. The details of the construction of the dataset, including the anonymization and pseudonymization process, are available in Appendix~\ref{app:mm_dataset}.

A validation set composed of 4096 studies was excluded from the training set and used to adapt the learning rate and stop training on plateau. The training set was then composed of 215,579 studies.

The training was executed on a single Nvidia A100 80GB GPU using PyTorch 2 with fp16 mixed precision. A batch size of 96 was employed for the initial 224x224 resolution, alongside a LION~\cite{chen2023symbolic} optimizer quantized in 8bits~\cite{DBLP:journals/corr/abs-2110-02861} with a learning rate of $1e^{-5}$, which was reduced by a factor of 2 following a plateau of 3 epochs of validation loss. Additionally, a weight decay of $1e^{-5}$ was applied. Training ceased after 10 epochs without any improvement in validation loss, and the model exhibiting the best validation loss was retained.
The training with the 336x336 and 448x448 resolutions was restarted from the training described above with a batch size kept as 96 for the ViT B24 336 and B32 448, but was reduced to 64 for the ViT B16 336 and to 48 for the ViT B16 448 models due to increased memory requirements. A learning rate of $1e^{-6}$ was used for this second training phase with the same learning rate scheduling and stopping strategy as the first phase.

As data augmentations, a random resized crop to 512x512 followed by a normalisation (mean 0.5, std 0.25), an horizontal flip ($p=0.5$), an affine transformation (random rotation from $-20$ to $+20^\circ$ and translation from $-10$ to $+10\%$), brightness and contrast adjustment (random from $0.8$ to $1.2$ for both), Gaussian blur (random sigma from $0.1$ to $3.0$), and final resizing to 224x224, 336x336 or 448x448 depending on the image encoder resolution.

Performance evaluations were carried out on the downstream tasks specified in Section \ref{sec:tasks_exp}.

\subsection{Evaluation on downstream tasks}\label{sec:tasks_exp}

It is crucial to note that our objective is not to reach state of the art performance for each dataset but rather to assess the effectiveness of the pretraining process. With this objective in mind, no data augmentation was conducted beyond normalization and resizing to the target resolution. In addition to an evaluation on the validation dataset used in Section~\ref{sec:exp:supervised_validation}, 5 more dataset are used for these evaluations, they are detailed in Appendix~\ref{app:down_datasets} and summarized in Table~\ref{tab:datasets}.

\subsubsection{Classification on images from the hospital considered by VLP}\label{sec:exp:supervised_validation}

\begin{figure}[h!]
    \centering
    \includegraphics[width=0.8\textwidth]{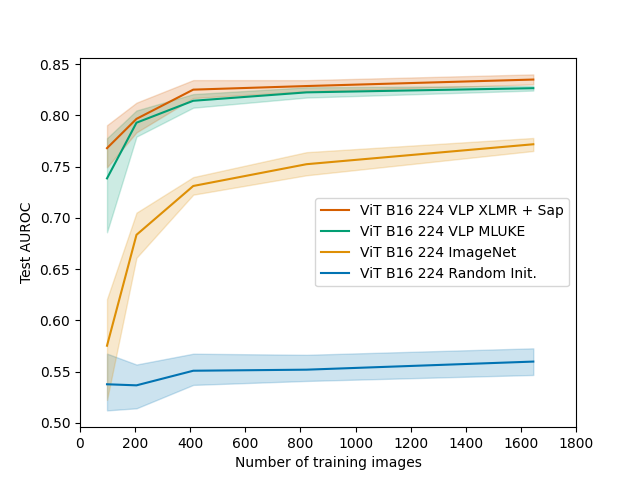}
    \caption{Classification AUROC achieved when training a \textbf{linear projection of a frozen vision encoder} on varying numbers of images obtained from the same hospital as the dataset used for vision language pretraining (VLP). The shaded areas around the lines represent the 95\% confidence intervals, calculated from 8 training sessions, with different seeds used for sampling these images. The VLP models achieved better performance than a model trained on ImageNet, even when using an order of magnitude fewer images during training. Only two VLP models are represented, the others are shown in Appendix~\ref{app:osrx_vlp_all}.}
    \label{fig:osrx_plot}
\end{figure}

The validation dataset is composed of 1351 studies, of which 256 are reserved as a test set. See Appendix~\ref{app:val_dataset} for more details about the creation of this dataset. The remaining 1095 studies (accounting for 3657 X-Rays) are randomly sampled as a validation set using 10\% of the studies, and a train set using a subset with a ratio of 0.5, 0.25, 0.125, 0.0625 and 0.03 of the remaining data. The effective train set size is thus comprised between 30 and 493 studies ($\approx$ 99 to 1646 X-Rays).

A linear layer has been added to the frozen pre-trained image encoder, to be trained with a binary cross-entropy loss function, weighted by the ratio between positives and negatives in the training set. The initial learning rate was set to 1e-4 and halved after 3 epochs without a decrease in validation loss. Training ceased after 10 epochs without improvement in validation loss, with the best-performing model retained for evaluation on the test set.

The results are represented in Figure~\ref{fig:osrx_plot}.

The VLP models achieved significantly better performance than a model trained on ImageNet, even when using one order of magnitude fewer images during training.

\begin{table}[]
    \centering
    \begin{tabular}{ll|ccccc}
          &                & MURA & FracAtlas & OAI KL & OAI HKA & RSNA Bone Age \\
          \toprule
\multirow{2}{4em}{Zero-shot} & Retrieval      & \checkmark  & \checkmark & - & - & - \\
          & Classification & \checkmark  & \checkmark & - & - & - \\
          \midrule
\multirow{2}{4em}{Trained}   & Classification & \checkmark  & \checkmark & \checkmark & - & - \\
          & Regression     & -  & - & - & \checkmark & \checkmark
    \end{tabular}
    \caption{Downstream tasks datasets. A more extended description is available in Appendix~\ref{app:down_datasets}.}
    \label{tab:datasets}
\end{table}

\subsubsection{Trained classification on images outside the VLP hospital}\label{sec:trained_classif}

\begin{table}[h]
    \centering
    \begin{tabular}{lcccccccr}
        \toprule
         & \multicolumn{2}{c}{FracAtlas} & \multicolumn{3}{c}{MURA} & \multicolumn{2}{c}{OAI KL scale} & Mean difference\\
         & \multicolumn{2}{c}{(AUROC)} & \multicolumn{3}{c}{(AUROC)} & \multicolumn{2}{c}{(AUROC)} & to ImageNet \\
        Train set ratio & 10\% & all & 1\% & 10\% & all & 10\% & all & ($\Delta$ AUROC)\\
        \midrule
        \multicolumn{8}{l}{\textit{A. General initialization methods (ViT B16 224)}} \\
        ImageNet Init. & 78.6 & 86.8 & 70.1 & 81.1 & 83.0 & 58.6 & 67.8 & 0 \\
        Random Init. & 62.6 & 66.2 & 56.7 & 57.4 & 58.7 & 51.9 & 53.8 & - 15.7 \\
        \midrule
        \multicolumn{8}{l}{\textit{B. English based VLP (ResNet50, results from the paper~\cite{Zhang2022convirt})}} \\
        ConVirt & - & - & 81.2 & 85.1 & 87.6 & - & - & - \\
        \midrule
        \multicolumn{8}{l}{\textit{C. Our French based VLP - Text encoder (+ ViT B16 224)}} \\
        Dr BERT & 87.3 & 89.8 & 81.2 & 84.4 & 86.2 & 68.7 & 71.7 & + 6.2 \\
        Dr BERT + Sap & 89.3 & 90.8 & 82.2 & 84.9 & 86.6 & 70.5 & 73.0 & + 7.3 \\
        MLUKE & 88.8 & 91.8 & 80.3 & 84.7 & 86.4 & 67.1 & 70.5 & + 6.2 \\
        MLUKE + Sap & 90.5 & 92.8 & 82.2 & 84.8 & 86.9 & 68.7 & 71.6 & + 7.4 \\
        XLMR & 88.7 & 91.5 & 80.5 & 84.3 & 86.0 & 68.1 & 71.5 & + 6.4 \\
        XLMR + Sap & 88.2 & 91.0 & 83.2 & 85.7 & 87.0 & 69.8 & 72.4 & + 7.3 \\
        \midrule
        \multicolumn{8}{l}{\textit{D. ViT resolution increases to 336x336 from C.}} \\
        ViT B16 XLMR + Sap & \textbf{92.2} & \textbf{94.4} & 85.1 & \textbf{87.1} & 88.3 & 74.3 & \textbf{76.2} & \textbf{+ 10.2} \\
        ViT B16 mluke & 90.2 & 93.1 & 84.9 & 86.4 & 87.8 & 71.9 & 74.5 & + 9.0\\
        ViT B24 XLMR + Sap & 89.7 & 92.3 & 83.9 & 86.1 & 87.5 & 70.3 & 72.8 & + 8.1 \\ 
        ViT B24 mluke & 89.4 & 92.6 & 83.0 & 85.5 & 87.2 & 69.2 & 72.1 & + 7.5 \\
        \midrule
        \multicolumn{7}{l}{\textit{D. ViT resolution increases to 448x448 from C.}} \\
        ViT B16 XLMR + Sap & 91.2 & 94.2 & \textbf{85.6} & 86.7 & \textbf{88.4} & \textbf{74.5} & \textbf{76.2} & + 10.1 \\
        ViT B16 mluke & 89.8 & 93.0 & 84.7 & 86.9 & 88.3 & 73.0 & 75.2 & + 9.3 \\
        ViT B32 XLMR + Sap & 89.2 & 92.5 & 83.6 & 86.0 & 87.4 & 70.1 & 72.6 & + 7.9\\
        ViT B32 mluke & 89.2 & 92.2 & 83.6 & 85.7 & 87.3 & 69.4 & 72.2 & + 7.7 \\
        \bottomrule
    \end{tabular}
    \vspace{0.2cm}
    \caption {Classification results obtained with \textbf{linear projection of a frozen vision encoder}, for the text encoders presented in Section \ref{sec:text_encoder}, with our without text synonym self-alignment (Sap). Different training set ratios are considered to evaluate how the amount of training samples impacts the benefit obtained from VLP pretraining. All vision-language pretrained (VLP) models performs better than ImageNet ViT, with less data. Our models trained with French reports are also on par with ConVirt for MURA dataset at 224x224 resolution. Resolution increase improves the results.}
    \label{tab:classif_lin}
\end{table}

\begin{table}[h]
    \centering
    \begin{tabular}{lcccccccr}
        \toprule
         & \multicolumn{2}{c}{FracAtlas} & \multicolumn{3}{c}{MURA} & \multicolumn{2}{r}{OAI KL scale} & Mean difference\\
         & \multicolumn{2}{c}{(AUROC)} & \multicolumn{3}{c}{(AUROC)} & \multicolumn{2}{r}{(AUROC)} & to ImageNet \\
        Train set ratio & 10\% & all & 1\% & 10\% & all  & 10\% & all & ($\Delta$ AUROC)\\
        \midrule
        \multicolumn{8}{l}{\textit{A. General initialization methods (ViT B16 224)}} \\        ImageNet Init. & 80.3 & 88.6 & 70.3 & 81.8 & 87.0 & 66.1 & 75.3 & 0\\
        Random Init. & 66.6 & 69.0 & 57.5 & 60.0 & 64.9 & 51.8 & 54.1 & - 17.9\\
        \midrule
        \multicolumn{8}{l}{\textit{B. English based VLP (ResNet50, results from the paper~\cite{Zhang2022convirt})}} \\
        ConVirt & - & - & 81.3 & 86.5 & 89.0 & - & - & - \\
        \midrule
        \multicolumn{8}{l}{\textit{C. Our French based VLP - Text encoder (+ ViT B16 224)}} \\
        Dr BERT & 88.5 & 91.4 & 81.2 & 85.8 & 89.5 & 71.6 & 77.6 & + 5.2 \\
        Dr BERT + Sap & 89.5 & 93.2 & 82.0 & 85.0 & 88.9 & 72.4 & 78.1 & + 5.7 \\
        MLUKE & 88.9 & 92.4 & 78.4 & 84.3 & 88.8 & 71.2 & 78.1 & + 4.7 \\
        MLUKE + Sap & 89.3 & 93.2 & 82.0 & 86.0 & 89.6 & 71.5 & 77.0 & + 5.6 \\
        XLMR & 89.7 & 93.9 & 79.9 & 84.1 & 89.6 & 71.9 & 77.6 & + 5.3 \\
        XLMR + Sap & 89.9 & 92.4 & 82.7 & 85.9 & 89.3 & 71.6 & 78.1 & + 5.8 \\
        \midrule
        \multicolumn{7}{l}{\textit{D. ViT resolution increases to 336x336 from C.}} \\
        ViT B16 XLMR + Sap  & \textbf{93.3} & \textbf{96.1} & 84.3 & \textbf{88.0} & \textbf{90.8} & 74.7 & 82.0 & \textbf{+ 8.5} \\
        ViT B16 MLUKE & 90.5 & 93.3 & 82.3 & 86.7 & 90.2 & 74.5 & 80.2 & + 6.9 \\
        ViT B24 XLMR + Sap & 90.3 & 93.3 & 82.5 & 86.3 & 89.5 & 72.0 & 78.5 & + 6.1 \\
        ViT B24 MLUKE & 89.6 & 93.5 & 80.3 & 85.6 & 89.5 & 71.6 & 77.5 & + 5.4\\
        \midrule
        \multicolumn{7}{l}{\textit{D. ViT resolution increases to 448x448 from C.}} \\
        ViT B16 XLMR + Sap & 91.3 & 95.3 & \textbf{84.4} & 87.6 & 90.5 & \textbf{75.3} & \textbf{82.3} & + 8.2\\
        ViT B16 mluke & 90.6 & 94.1 & 82.2 & 86.8 & 90.5 & 74.7 & 80.2 & + 7.1\\
        ViT B32 XLMR + Sap & 91.2 & 93.0 & 82.7 & 85.9 & 89.4 & 71.8 & 78.7 & + 6.2\\
        ViT B32 mluke & 89.3 & 93.3 & 80.8 & 85.6 & 89.3 & 71.6 & 77.7 & + 5.4\\
        \bottomrule
    \end{tabular}
    \vspace{0.2cm}
    \caption{Classification performance obtained with \textbf{fine-tuning of the vision encoder}, followed by a linear projection, for the text encoders presented in Section \ref{sec:text_encoder}, with our without text synonym self-alignment (Sap). Different training set ratios are considered to evaluate how the amount of training samples impacts the benefit obtained from VLP pretraining. All VLP models performs better than ImageNet ViT with less data. Our models trained with French reports are also on part with ConVirt for MURA dataset at 224x224 resolution. Resolution increase improves the results.}
    \label{tab:classif_ft}
\end{table}

A linear layer has been added to the pre-trained image encoder, to be trained with a binary cross-entropy loss function, weighted by the ratio between positives and negatives in the training set. The initial learning rate was set to 1e-4 and halved after 3 epochs without a decrease in validation loss. Training ceased after 10 epochs without improvement in validation loss, with the best-performing model retained for evaluation on the test set.

Two training scenarios are considered. In the first one, the vision encoder is frozen, and only the linear projection layer is updated during training. In the second one, after having been frozen for 200 steps, the image encoder is unfrozen and fine tuned with a learning rate reduced to 1e-6 to mitigate rapid overfitting of the Vision Transformer (ViT) model.

The results for the trained classification task are shown in Table~\ref{tab:classif_lin} for the linear projection appended to frozen models, and in Table~\ref{tab:classif_ft} for the fine-tuning case. Different training set ratios are considered to evaluate how the amount of training samples impacts the benefit obtained from VLP pretraining. The results produced by our models always show superior performances in comparison to models initialised from ImageNet or from scratch, both in linear evaluation and fine-tuning. At a resolution of 224x224, our results are comparable or better than results from the ConVirt paper, where an English-based dataset of bone X-Rays was used. The comparison is limited since, although the settings are similar, they are not identical. Notably, the architecture of the image encoder and the pretraining datasets are different.

The Sap process tends to improve the classification results of the three encoders.
Among the different text encoders used as initialisation, all are performing similarly well on average when coupled with Sap for the 224x224 resolution comparison, with small variations on the individual datasets.

The increase of resolution during the vision-language pretraining from 224x224 to 336x336 has a positive impact on the performances. Keeping the patch size of 16x16 while increasing the resolution by interpolating the position embedding requires more computations but shows better performances in comparison to increasing the patch size to 24x24, allowing to keep a complexity similar to a ViT B16 on 224x224 images.

Further increasing the resolution to 448x448 does not improve significantly the results, and is often detrimental in comparison to a 336x336 resolution. A possible explanation for this phenomenon could be related to the reduction of the batch size during vision-language pretraining for the ViT B16 448 models.

\subsubsection{Regression}

\begin{table}[h!]
    \centering
        \begin{tabular}{lrrr|rrrr}
        \toprule
        dataset & \multicolumn{3}{c}{OAI HKA} & \multicolumn{4}{c}{RSNA bone age} \\
         & \multicolumn{3}{c}{(Mean error in °)} & \multicolumn{4}{c}{(Mean error in months)} \\
        eval & lin. 10\% & lin. all & ft all & lin 10\% & lin all & ft 10\% & ft all \\
        \midrule
        \multicolumn{7}{l}{\textit{A. General initialization methods (ViT B16 224)}} \\
        ImageNet Init. & \textbf{2.42} & \textbf{2.04} & 1.68 & 16.20 & 14.99 & 15.56 & 12.08 \\
        Random Init. & 2.68 & 2.66 & 2.61 & 32.08 & 31.35 & 32.37 & 22.71 \\
        \midrule
        \multicolumn{7}{l}{\textit{C. Our French based VLP - Text encoder (+ ViT B16 224)}} \\
        Dr BERT & 2.62 & 2.33 & 1.64 & 16.43 & 15.37 & 14.91 & 11.45 \\
        Dr BERT + Sap & 2.58 & 2.24 & 1.63 & 15.85 & 14.54 & 15.16 & 11.39 \\
        MLUKE & 2.64 & 2.37 & 1.58 & 16.56 & 15.51 & 14.99 & 11.34 \\
        MLUKE + Sap & 2.56 & 2.26 & 1.69 & 15.86 & 14.71 & 15.00 & 11.55 \\
        XLMR & 2.56 & 2.26 & 1.62 & 16.34 & 15.29 & 15.32 & 11.59 \\
        XLMR + Sap & 2.64 & 2.40 & 1.56 & 15.50 & 14.46 & 14.78 & 11.41 \\
        \midrule
        \multicolumn{7}{l}{\textit{D. ViT resolution increases to 336x336 from C.}} \\
        ViT B16 MLUKE & 2.60 & 2.33 & 1.51 & 14.86 & 13.96 & 14.18 & 10.74 \\
        ViT B16 XLMR + Sap & 2.56 & 2.27 & 1.56 & 14.73 & 13.77 & 14.23 & 10.88 \\
        ViT B24 MLUKE & 2.66 & 2.31 & 1.57 & 15.35 & 14.36 & 14.56 & 11.28 \\
        ViT B24 XLMR + Sap & 2.58 & 2.31 & 1.60 & 15.51 & 14.65 & 14.98 & 11.15 \\
        \midrule
        \multicolumn{7}{l}{\textit{D. ViT resolution increases to 448x448 from C.}} \\
        ViT B16 MLUKE & 2.64 & 2.27 & \textbf{1.47} & 14.72 & 13.93 & 13.56 & 10.28 \\
        ViT B16 XLMR + Sap & 2.60 & 2.31 & 1.54 & \textbf{14.43} & \textbf{13.41} & \textbf{13.51} & \textbf{10.15} \\
        ViT B32 MLUKE & 2.60 & 2.26 & 1.55 & 15.44 & 14.51 & 14.72 & 11.35 \\
        ViT B32 XLMR + Sap & 2.64 & 2.36 & 1.58 & 15.41 & 14.49 & 15.04 & 11.26 \\

        \bottomrule
        \end{tabular}
    \vspace{0.2cm}
    \caption{Regression tasks. Two training scenarios are envisioned:  linear layer training with frozen encoder (lin.), and entire network fine-tuning (ft). Our VLP models do not perform better than ImageNet for angle measurement, probably due to scale invariant pretraining. The Bone age estimation resulting from the linear projection is improved by the VLP pretraining when the full model is fine-tuned, but not with fixed encoder. Resolution increases performances for most scenarios.}
    \label{tab:regression}
\end{table}

In the regression context, a single linear layer is appended to the CLS token of the image encoder, with additional scale and bias parameters initialized using the mean and standard deviation of the training set for each dataset. To train the resulting model, a smooth L1 loss function~\cite{Huber1964RobustEO} is used, and the mean absolute deviation (MAD) serves as the test prediction evaluation metric.

 In practice, as for the classification case, two training scenarios are considered. The first one keeps the encoder frozen, while the second fine tunes it.

For the RSNA Pediatric Bone Age dataset, the linear layer takes as an additional input the sex of the patient.
We compare the effectiveness of training with either the entire training set or 10\% of both the RSNA Pediatric Bone Age and OAI HKA Angles datasets. As with the classification task, data augmentation is minimized, involving only resizing to the target resolution and normalization with a mean of 0.5 and a standard deviation of 0.25. For the OAI HKA measurement exclusively, resizing was conducted while preserving the aspect ratio by padding the image. This approach aimed to prevent distortion of the angles within the image.



The results of the regression task can be seen in Table~\ref{tab:regression}.

For the RNSA bone age estimation, our pretraining is beneficial in all fine-tuning cases in comparison to an ImageNet or random initialisation, and with all or 10\% of the training set.
The increase of resolution is also beneficial, but only when we scale the number of token by interpolating the position embeddings and keeping the 16x16 patch size. Increasing the resolution by changing the patch size does not yield better results.

Concerning the OAI HKA measurement, the error rate is higher than the bacis ImageNet model in linear evaluation, and only a full fine-tuning allows to produce similar performances on 224x224 images. The increase of resolution has no effect for the linear evaluation, and slightly increases the performances for the fine-tuned models. These results are not surprising given that the aspect ratio of the images is modified during the data augmentation of our vision-language pretraining. This makes our models invariant to this kind of deformations but also impedes the ability to measure angles. 

\subsubsection{Zero-Shot Classification}\label{sec:zero_classif}

\begin{table}[!h]
    \centering
    \begin{tabular}{lcccc}
    \toprule
       & Text  & Text & Latent & Latent \\
    & Binary & Enumeration & Minimum & Mean \\
    \midrule
    \multicolumn{5}{l}{\textit{A. VLP pretraining - Text encoder (+ ViT B16 224)}} \\
    Dr BERT & 67.4 & 66.8 & 64.8 & 65.6 \\
    Dr BERT + Sap & 69.2 & 76.9 & 68.8 & 76.8 \\
    MLUKE & 73.9 & 74.7 & 72.3 & 74.6 \\
    MLUKE + Sap & 68.0 & 72.5 & 67.0 & 68.9 \\
    XLMR & 60.3 & 72.6 & 69.4 & 68.1 \\
    XLMR + Sap & 65.4 & 78.4 & 73.1 & 72.9 \\
    \midrule
    \multicolumn{5}{l}{\textit{B. VLP pretraining - 336x336 Image encoder \& Text encoder}} \\
    ViT B16 336 \& XLMR + Sap & 64.0 & \underline{\textbf{79.2}} & 74.4 & 74.2 \\
    ViT B24 336 \& XLMR + Sap & 64.0 & 78.1 & 72.7 & 73.0 \\
    ViT B16 336 \& MLUKE & \underline{75.9} & 75.9 & \underline{74.7} & \underline{77.4} \\
    ViT B24 336 \& MLUKE & 74.4 & 74.3 & 73.7 & 75.0 \\
    \midrule
    \multicolumn{5}{l}{\textit{C. VLP pretraining - 448x448 Image encoder \& Text encoder}} \\
    ViT B16 448 \& XLMR + Sap & 62.7 & \underline{\textbf{79.2}} & 73.4 & 74.0 \\
    ViT B32 448 \& XLMR + Sap & 65.5 & 78.5 & 72.2 & 72.4 \\
    ViT B16 448 \& MLUKE & 74.6 & 74.5 & 73.5 & 77.0 \\
    ViT B32 448 \& MLUKE & 73.9 & 74.1 & 73.5 & 75.0 \\
    
    \bottomrule
    \end{tabular}
    \vspace{0.2cm}
    \caption{\centering
    \textbf{Zero-Shot classification on MURA}
    with different image and text encoders. Sap denotes the use of textual synonyms self-alignment.
    Four strategies have been considered to localize normal/abnormal classes in the embedding space, from text prompts. 'Text binary' simply uses the name of the classes (normal/abnormal) as text prompts. 'Text enumeration' uses a prompt consisting of a list of pathologies (see text for details) separated by commas for the abnormal class. 'Latent minimum' uses individual embeddings for each sub-classes. Eventually, 'latent mean' averages, in the embedding space, the prompts derived from each pathology associated to the abnormal class.
    Best overall in \textbf{Bold}, best for each strategy in \underline{underline}. MLUKE performs best with latent mean strategy while XLMR + Sap is best with text enumeration. Performances increase with the 336x336 resolution, the 448x448 resolution does not improve.}
    \label{tab:zero_classif_mura}
\end{table}

\begin{table}[!h]
    \centering
        \begin{tabular}{lcccc}
        \toprule
         & Text & Text & Latent & Latent\\
         & Binary & Enumeration & Minimum & Mean \\
        \midrule
        \multicolumn{5}{l}{\textit{A. VLP pretraining - Text encoder (+ ViT B16 224)}} \\
        Dr BERT & 56.3 & 49.3 & 51.6 & 47.2 \\
        Dr BERT + Sap & 72.7 & 56.1 & 47.8 & 56.6 \\
        MLUKE & 72.8 & 62.0 & 67.3 & 66.9 \\
        MLUKE + Sap & 61.1 & 55.2 & 41.3 & 52.4 \\
        XLMR & 70.7 & 62.5 & 58.3 & 60.4 \\
        XLMR + Sap & 61.3 & 57.0 & 57.7 & 59.0 \\
        \midrule
        \multicolumn{5}{l}{\textit{B. VLP pretraining - 336x336 Image encoder \& Text encoder}} \\
        ViT B16 336 \& XLMR + Sap & 71.0 & \underline{66.2} & 65.8 & 68.6 \\
        ViT B24 336 \& XLMR + Sap & 62.6 & 59.1 & 60.8 & 61.1 \\
        ViT B16 336 \& MLUKE & \underline{\textbf{77.3}} & 56.6 & 63.9 & \underline{70.0} \\
        ViT B24 336 \& MLUKE & 72.4 & 58.2 & 66.3 & 67.4 \\
        \midrule
        \multicolumn{5}{l}{\textit{C. VLP pretraining - 448x448 Image encoder \& Text encoder}} \\
        ViT B16 448 \& XLMR + Sap & 69.2 & 63.6 & 64.5 & 66.5 \\
        ViT B32 448 \& XLMR + Sap & 61.0 & 57.8 & 57.5 & 58.4 \\
        ViT B16 448 \& MLUKE & 73.5 & 57.9 & \underline{66.7} & 69.9 \\
        ViT B32 448 \& MLUKE & 71.5 & 58.1 & 63.6 & 64.3 \\
        \bottomrule
        \end{tabular}
    \vspace{0.2cm}
    \caption{\centering
    \textbf{Zero-Shot classification on FracAtlas}  with different image and text encoders. SAP denotes the use of textual synonyms self-alignment.
    Four strategies have been considered to localize normal/abnormal classes in the embedding space, from text prompts. 'Text binary' simply uses the name of the classes as text prompts. 'Text enumeration' uses a prompt consisting of a list of pathologies separated by commas for the abnormal class. 'Latent minimum' uses individual embeddings for each sub-classes. Eventually, 'latent mean' averages, in the embedding space, the prompts derived from each pathology associated to the abnormal class.
    Best overall in \textbf{Bold}, best for each strategy in \underline{underline}. Similarly to Table~\ref{tab:zero_classif_mura}, the couples MLUKE with binary strategy and XLMR + Sap with text enumeration are the best performers. The 336x336 resolution shows superior performance with no benefits to further increase to 448x448.}
    \label{tab:zero_classif_fracatlas}
\end{table}

This task was explored using MURA and FracAtlas datasets. We chose not to pursue exploration on the OAI KL dataset for this task. This is because, unlike distinctive classes such as the presence or absence of bone fractures, the semi-quantitative KL scale poses a bigger challenges in being accurately reflected through text prompts.

Regarding the MURA dataset, the class prediction associated to a study is generated by averaging the results for all images within a given study. This dataset poses a challenge for zero-shot binary classification between normal and abnormal due to contextual variability in defining abnormality. For instance, the presence of osteoarthritis in a radiograph taken for an elderly individual following trauma to rule out a bone fracture or dislocation could be noted, yet the overall radiograph might still be treated as normal within the trauma context.

Therefore, we explored several strategies, as described in Section~\ref{sec:zero_tasks}, to localize the normal and abnormal classes in the shared embedding space. 

The details about the exact prompts explored are described in Appendix~\ref{app:zero_strategies} for both MURA and FracAtlas.

The methodology described in Section~\ref{sec:zero_tasks} was evaluated and reported on the test set of both datasets. Evaluation of this task is conducted using the Area under the ROC curve (AUROC).

Results are presented in Table~\ref{tab:zero_classif_mura} and Table~\ref{tab:zero_classif_fracatlas} for MURA and FracAtlas, respectively.

These results exhibit significant variations in performance depending on the prompting strategy and on the chosen models. This is not surprising since we employ a straightforward distance measurement between image embeddings and anchor points that differ substantially from conventional radiology reports. 

Depending on the models, a text enumerating various pathologies considered as abnormal or the average of the embeddings of these pathologies performs best on MURA, while a simpler text query performs better for the binary bone fracture classification of FracAtlas. For both datasets, encoders based on MLUKE and XLMR + Sap performs best. The overall effect of the Sap pretraining is not clearly demonstrated as beneficial in this experiment. For MLUKE, this degrades systematically the performances.

Increasing the resolution from 224x224 to 336x336 improves results for both setups explored, while further increase to 448x448 does not improve results.

Our analysis reveals that the models exhibit significant sensitivity to the prompt employed, raising the possibility that alternative prompting strategies may yield improved outcomes for one model or another. Consequently, it remains challenging to definitively determine which encoder is optimal on this task.

\subsubsection{Zero-shot Text-Image Retrieval}

In this experiment, the MURA and FracAtlas datasets were also used. Instead of evaluating only using the test set, as decided for the classifications tasks (to allow comparison with previous works from ConVirt), we split each dataset in 5 folds and performed the retrieval task on each fold. The retrieval precision was computed on the top k retrieved images with k=10 and 50 and without any training on theses datasets.

Results are represented in Table~\ref{tab:fracatlas_retrieval} and Table~\ref{tab:mura_retrieval} for MURA and FracAtlas, respectively.

For MURA, the differences between models are smaller in comparison to zero-shot classification. The text synonyms self-alignment pretraining (Sap) increases performances of the Dr BERT model on FracActlas, while degrading performances of the other models. On MURA however, the difference between models with or without Sap is smaller and not significantly different.

\begin{table}[!h]
    \centering
    \begin{tabular}{l|cc|cc|cc}
        \toprule
 ViT B16 224 \& & \multicolumn{2}{c|}{Negation} & \multicolumn{2}{c|}{Text enumeration} & \multicolumn{2}{c}{Latent mean} \\
 Text encoder& Prec@10 & Prec@50 & Prec@10 & Prec@50 & Prec@10 & Prec@50 \\
\midrule
Dr BERT & 64 (12.9) & 56.0 (5.4) & 51.0 (7.4) & 48.6 (4.4) & 52 (4.5) & 51.2 (4.3) \\
Dr BERT + Sap & 90 (6.1) & 82.8 (5.8) & 75 (15.0) & 64.0 (4.8) & 63 (9.7) & 62.6 (5.9) \\
XLMR & 85 (6.1) & 77.2 (3.3) & 89 (12.4) & 68.4 (6.6) & 76 (8.2) & 64.2 (4.9) \\
XLMR + Sap & 68 (10.4) & 65.2 (4.6) & 60 (6.1) & 60.2 (2.4) & 69 (8.9) & 66.8 (5.4) \\
MLUKE & 95 (3.5) & 78.4 (1.9) & 77 (6.7) & 64.8 (2.9) & 80 (6.1) & 72.4 (2.1) \\
MLUKE + Sap & 88 (7.6) & 71.6 (2.9) & 75 (8.7) & 63.4 (4.9) & 74 (6.5) & 59.2 (4.9) \\
        \bottomrule
    \end{tabular}
    \caption{FracAtlas Retrieval Results. Prec@X denotes the retrieval precision among the top X samples. (standard deviation in parenthesis, computed using a 5-fold). In this experiment, results show more variability, with model performances diverging from those in the other experiments. XLMR shows superior performance than XLMR + Sap, while Dr BERT + Sap demonstrates good performance. Consistently with previous experiments, MLUKE remains among the top performers, while Dr BERT alone consistently ranks at the bottom.
    }
    \label{tab:fracatlas_retrieval}
\end{table}

\begin{table}[!h]
    \centering
    \begin{tabular}{l|rr|rr|rr}
        \toprule
         ViT B16 224 \& & \multicolumn{2}{c|}{Negation} & \multicolumn{2}{c|}{Text enumeration} & \multicolumn{2}{c}{Latent mean} \\
          Text encoder & Prec@10 & Prec@50 & Prec@10 & Prec@50 & Prec@10 & Prec@50 \\
        \midrule
        Dr BERT & 86.0 (8.2) & 86.4 (4.0) & 93.0 (5.7) & 86.6 (2.9) & 85.0 (5.0) & 84.4 (2.9) \\
        Dr BERT + Sap & 86.0 (4.2) & 86.2 (2.3) & 89.0 (4.2) & 89.8 (4.9) & 87.0 (8.4) & 90.4 (3.4) \\
        XLMR & 85.0 (3.5) & 87.0 (3.2) & 80.0 (6.1) & 83.2 (1.9) & 79.0 (4.2) & 81.2 (2.7) \\
        XLMR + Sap & 84.0 (6.5) & 84.6 (3.0) & 84.0 (2.2) & 87.2 (1.1) & 81.0 (11.4) & 84.8 (2.6) \\
        MLUKE & 89.0 (4.2) & 87.8 (2.3) & 89.0 (5.5) & 87.2 (2.6) & 90.0 (3.5) & 91.2 (1.3) \\
        MLUKE + Sap & 91.0 (5.5) & 85.8 (4.0) & 84.0 (9.6) & 85.8 (1.8) & 78.0 (4.5) & 80.8 (3.5) \\
        \bottomrule
    \end{tabular}
    \caption{MURA retrieval (standard deviation in parenthesis computed using a 5-fold). All models seems to perform well on the retrieval task for the MURA dataset, in contrast to Fracatlas in Table~\ref{tab:fracatlas_retrieval}. No significant difference can be consistently found between models.}
    \label{tab:mura_retrieval}
\end{table}

\subsection{Latent space exploration}\label{sec:latent}

In this section, we will explore in more details the native latent space organization of our models. The goal is to enhance our understanding of the results obtained in Section~\ref{sec:tasks_exp}. Particularly to get an insight for the zero-shot results variability.

\begin{figure}[!h]
    \centering
    \begin{subfigure}{0.32\linewidth}
        \includegraphics[width=\linewidth]{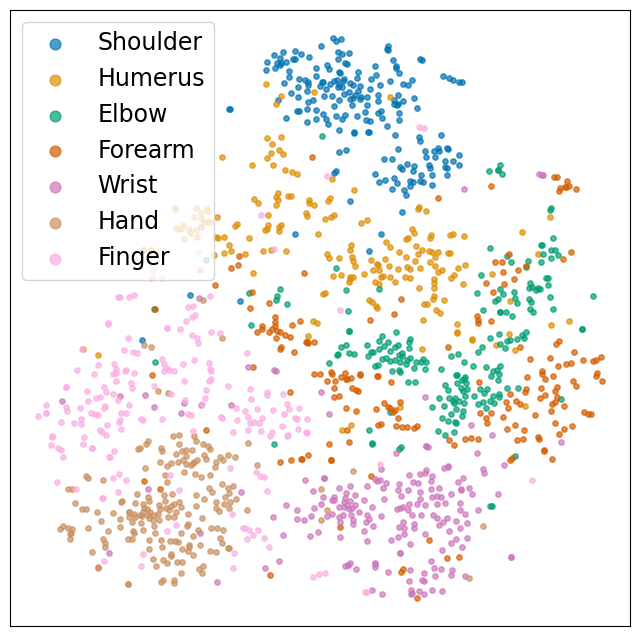}
        \caption{no VLP (ImageNet), locations}
        \label{fig:tsne_mura:loc_in}
    \end{subfigure}
    \begin{subfigure}{0.32\linewidth}
        \includegraphics[width=\linewidth]{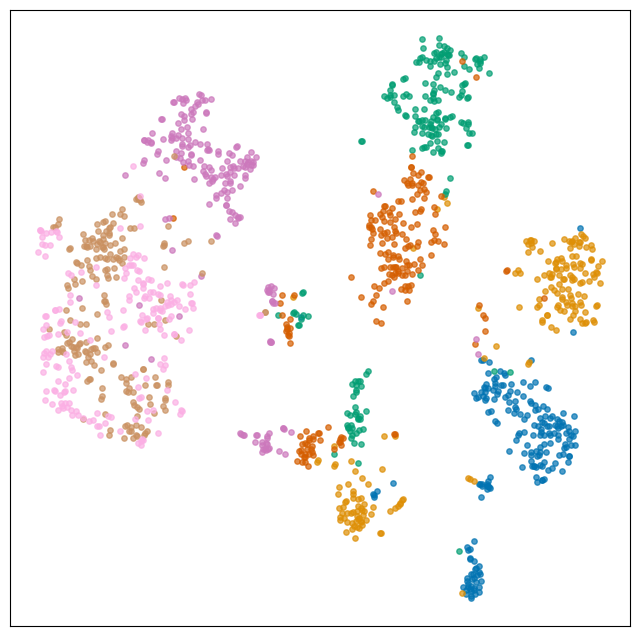}
        \caption{VLP with MLUKE, locations}
        \label{fig:tsne_mura:loc_mluke}
    \end{subfigure}
    \begin{subfigure}{0.32\linewidth}
        \includegraphics[width=\linewidth]{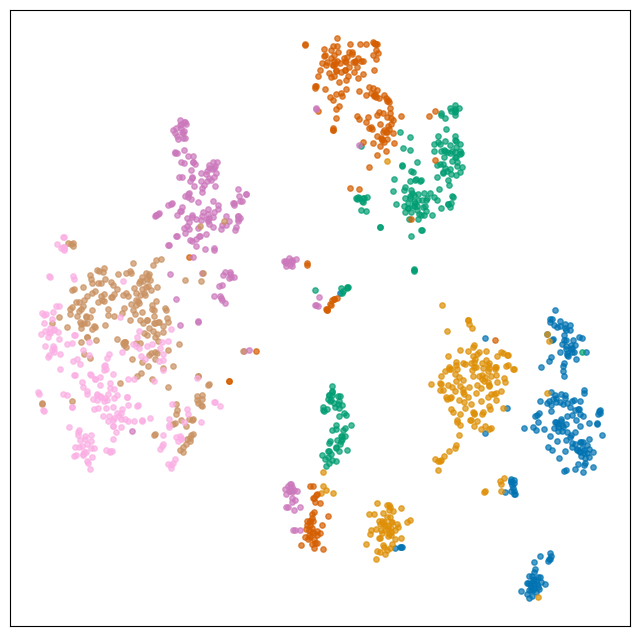}
        \caption{VLP with XLMR+Sap, locations}
        \label{fig:tsne_mura:loc_xlmrsap}
    \end{subfigure}
    \begin{subfigure}{0.32\linewidth}
        \includegraphics[width=\linewidth]{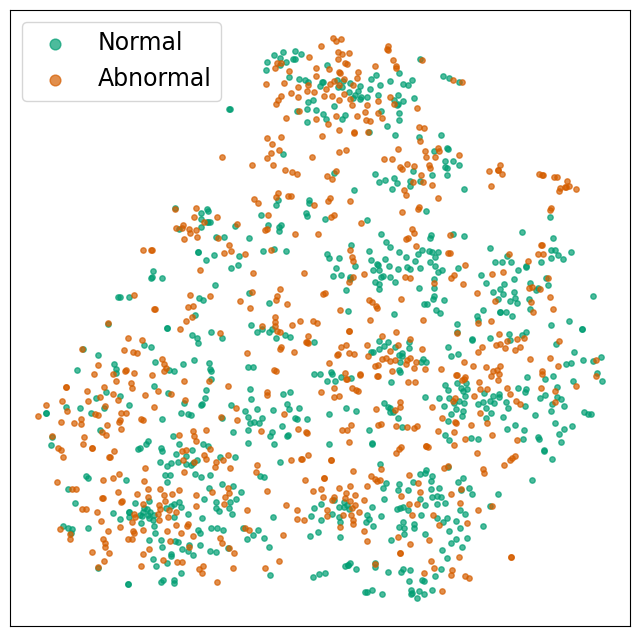}
        \caption{no VLP (ImageNet), labels}
        \label{fig:tsne_mura:label_in}
    \end{subfigure}
    \begin{subfigure}{0.32\linewidth}
        \includegraphics[width=\linewidth]{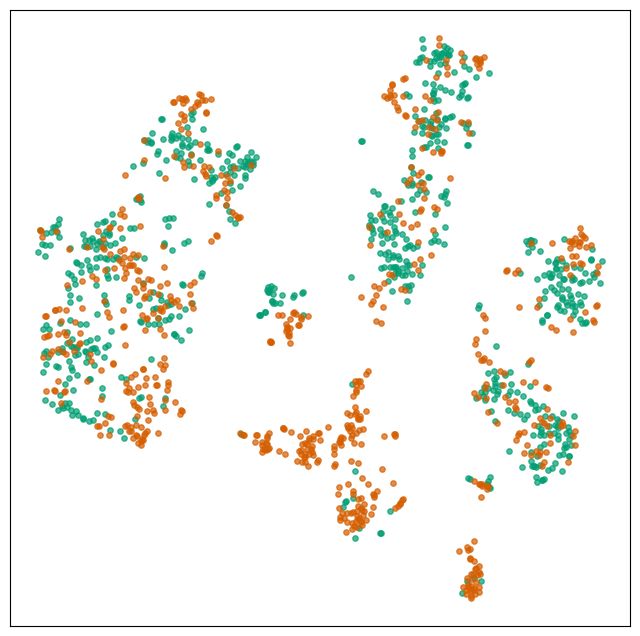}
        \caption{VLP with MLUKE, labels}
        \label{fig:tsne_mura:label_mluke}
    \end{subfigure}
    \begin{subfigure}{0.32\linewidth}
        \includegraphics[width=\linewidth]{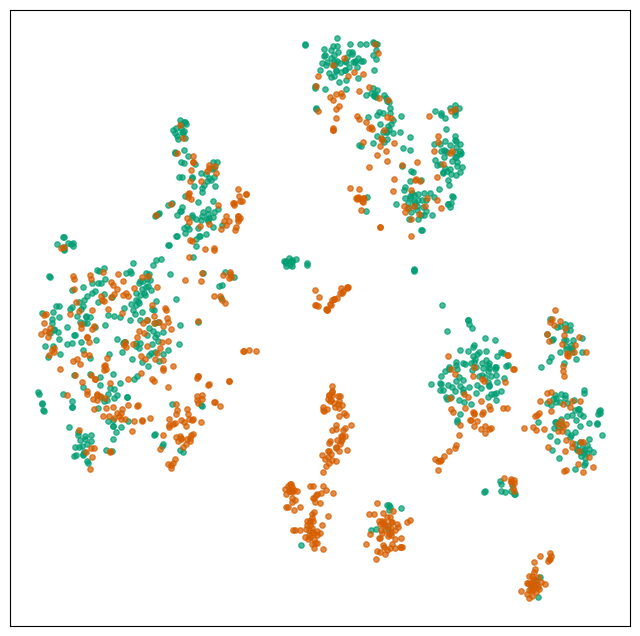}
        \caption{VLP with XLMR+Sap, labels}
        \label{fig:tsne_mura:label_xlmrsap}
    \end{subfigure}
    \caption{t-SNE visualizations of the embeddings of MURA images with and without VLP pretraining. The VLP models show a better clustering in comparison to the ImageNet model for both anatomical locations and labels. No training on MURA was conducted for any of the models. Clusters tends to form predominantly based on the anatomical location. However, within a specific anatomical site, various clusters frequently emerge, the most notable being clusters with osteosynthesis material (that can be visualized in the second line of the Figure as clusters composed of only abnormal images).
    Figures for all models can be found in Appendix~\ref{app:latent_additional}.}
    \label{fig:tsne_mura_loc}
\end{figure}

Using 200 images of each anatomical region of the MURA dataset and their associated embeddings produced by the self-supervised models, we employed a t-SNE~\cite{van2008visualizing} algorithm to explore visually intrinsic data distribution in the 512-dimensional space. The result can be seen in Figure~\ref{fig:tsne_mura_loc} for ImageNet, VLP pretrained with MLUKE, and VLP with XLMR + Sap. Notably, while a ViT B16 model trained on ImageNet has already begun to incompletely cluster the anatomical locations, a ViT models pretrained from Section~\ref{sec:vlp_pretraining} on bone X-Rays and French reports exhibits an improved ability to differentiate the different anatomical locations, with finger and hand being unsurprisingly the two locations with the higher overlap. The distribution of the normal/abnormal labels depicted in Figure~\ref{fig:tsne_mura:label_in}, \ref{fig:tsne_mura:label_mluke} and \ref{fig:tsne_mura:label_xlmrsap} is also interesting.

Two observations can be drawn from this analysis. Firstly, VLP models naturally form dense and well-separated clusters by anatomical region, unlike ImageNet. Secondly, for the VLP models, each anatomical region appears to be made up of large well-distinct sub-clusters and other smaller sub-clusters grouped together (as observed in the first row of the figure). These grouped smaller sub-clusters, corresponding to several anatomical regions, are drawn from pathological samples (as observed in the second row of the figure). By manually exploring the images belonging to different subclusters, we observe that osteosynthesis radiographs containing metal plates are more often grouped together and aggregated on separated groups on the t-SNE, in comparison to smaller or not displaced bone fractures that tend to remain closer to normal images of the same anatomical location in the t-SNE plot.

We fit a Linear Discriminant Analysis (LDA)~\cite{fisher1936use} using scikit-learn~\cite{scikit-learn} on the embeddings of the frozen models on the MURA training set and visualized the result on the test set as shown in Figure~\ref{fig:lda_mura}. The shifted peak observed in the LDA results for ours VLP models in comparison to ImageNet model indicates a better distinction between the two classes, making the classification task possible without fine tuning of the encoder. This is also consistent with results from the linear classification in Section~\ref{sec:trained_classif} and Table~\ref{tab:classif_lin} with ours VLP models performing better in lower data volume.

\begin{figure}[!h]
    \centering
    \begin{subfigure}{0.32\linewidth}
        \includegraphics[width=\linewidth]{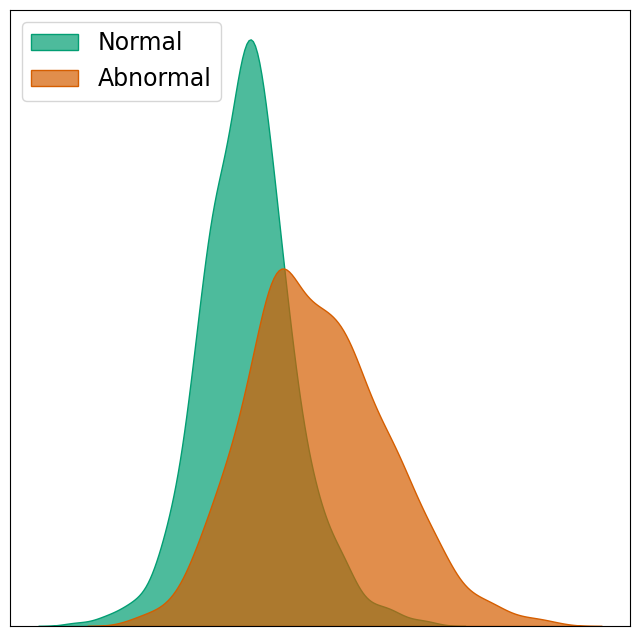}
        \caption{no VLP (ImageNet), labels}
    \end{subfigure}
    \begin{subfigure}{0.32\linewidth}
        \includegraphics[width=\linewidth]{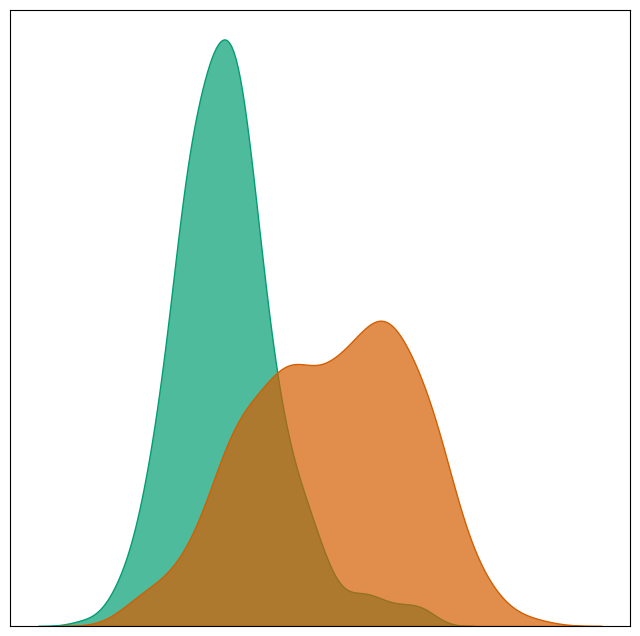}
        \caption{VLP with MLUKE, labels}
    \end{subfigure}
    \begin{subfigure}{0.32\linewidth}
        \includegraphics[width=\linewidth]{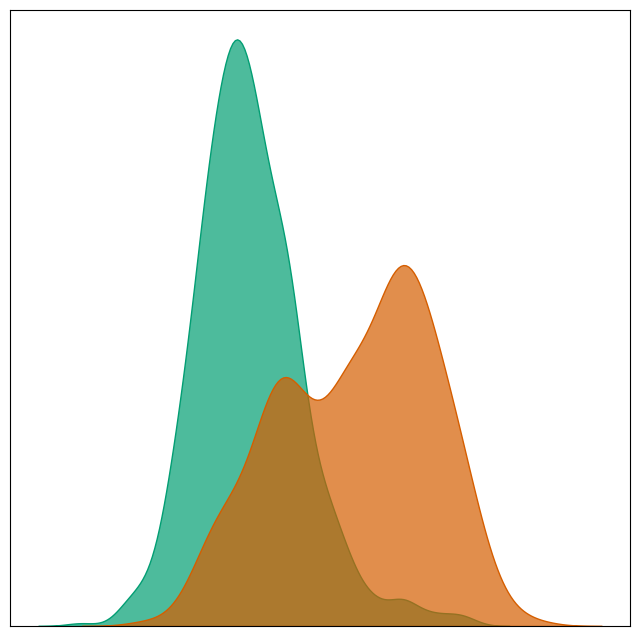}
        \caption{VLP with XLMR+Sap, labels}
    \end{subfigure}
    \caption{LDA visualizations of the embeddings of MURA images with and without VLP pretraining (not trained on MURA). The VLP models show a better separability in comparison to ImageNet with more shifted peaks. No training on MURA was conducted for any of the models. Figures for all models can be found in Appendix~\ref{app:latent_additional}.}
    \label{fig:lda_mura}
\end{figure}

\section{Conclusion}\label{sec:conclusion}

This paper has demonstrated the possibility of conducting contrastive vision-language pretraining using French reports related to bone radiographs collected in a single hospital. The whole pipeline to prepare the text and image data is made available with a special emphasis on the anonymization process adapted to French language.
We examined various text encoders initialisations and found that a multilingual text encoder outperforms those limited to biomedical French-only texts. Pretraining the text encoder through self-alignment using UMLS ontology has also improved performance on supervised downstream tasks.

In comparison to ImageNet, we observed a notable performance enhancement across different classification tasks, both when only training the final linear layer or when fine-tuning the whole network. There was also an improvement in regression tasks with fine-tuning of the model. While results in zero-shot settings are promising, they exhibit more variability, particularly in zero-shot classification, revealing the need for a minimal amount of annotations to solve tasks properly.

Increasing the resolution from 224x224 to 336x336 yielded better results, particularly when position embedding interpolation is used. However, further resolution increase to 448x448 showed limited or no additional gain, considering the increased computational complexity.


 Future research could involve incorporating a local contrastive loss, incorporating additional text reports such as orthopedic consultations or exploring more effective zero-shot and data-augmentation strategies.

\section*{Acknowledgments}
This work was funded by the Fond de la Recherche Scientifique de Belgique (F.R.S.-FNRS), including a FRIA (Fonds pour la formation à la Recherche dans l’Industrie et dans l’Agriculture) funding to Alexandre Englebert (ID 40014544).
Computational resources have been provided by the supercomputing facilities of the Université catholique de Louvain (CISM/UCL) and the Consortium des Équipements de Calcul Intensif en Fédération Wallonie Bruxelles (CÉCI) funded by the Fond de la Recherche Scientifique de Belgique (F.R.S.-FNRS) under convention 2.5020.11 and by the Walloon Region

\bibliographystyle{unsrt}  
\bibliography{references}

\pagebreak
\begin{center}
\textbf{\large Appendix}
\end{center}
\appendix

\section{Multi-modal Dataset for pretraining}\label{app:mm_dataset}

This section describes the creation of our dataset.
We obtained approval from the Hospital Ethics Committees (Belgian registration number B403201523492) to conduct this study, which involves the retrospective analysis of data from patients treated in the orthopedics department at Cliniques Universitaires Saint Luc in Brussels. 
A lot of effort was provided in order to protect the privacy of patients, following the GDPR\cite{GDPR}. This involved employing anonymization techniques when possible, and resorting to pseudonymization when complete anonymization was not feasible. Section \ref{sec:image_data} provides a breakdown of the procedures utilized in handling the images, Section \ref{sec:reports} delves into the processing of text and reports, and Section \ref{sec:pairing} explores how the two modalities are combined.

\subsection{Images preprocessing}\label{sec:image_data}

Our initial step involved identifying relevant patients by filtering the PACS (Picture Archiving and Communication System) to maintain patients who underwent imaging studies prescribed by the Orthopaedic surgeons of the hospital, and related to osteoarticular conditions. 
To ensure data anonymization, privacy related metadata were systematically removed and a new random unique identifier was assigned to each individual patient and to each study.

Upon manual examination of the images, it was observed that certain imaging devices included text reports with privacy-sensitive information (e.g., patient's name in dose reports). These reports were saved as images and mixed with X-rays. To address this issue, we employed the EasyOCR\footnote{\url{https://github.com/jaidedai/easyocr}} framework to extract text from the images with the objective of identifying problematic images. The choice of the framework was based on an initial comparison with Tesseract OCR\footnote{\url{https://github.com/tesseract-ocr/tesseract}}, our investigation indicated that the EasyOCR framework exhibited superior text identification capabilities when applied to our dataset. Subsequent manual inspection of the extracted texts revealed that images containing private patient information exhibited significantly more text than conventional X-ray images, which typically include simple indications such as laterality or patient position. We retained only those images with a character count below 35, a threshold we verified to be conservative. After this last filtering step, no residual patient privacy data was found in the retained images.

This final step yielded a dataset of 947,062 anonymized X-Ray images, grouped in 252,103 studies from 75,600 patients.

\subsection{Reports preprocessing}\label{sec:reports}

The process of extracting medical reports begins with identifying relevant documents in the Electronic Health Record (EHR) for patients previously identified in the PACS (Picture Archiving and Communication System) in Section \ref{sec:image_data}. A preliminary filtering step is implemented to select reports, focusing on medical analysis results (included radiology reports), consultation reports, hospitalization reports, and operation protocols.

Given that the reports are stored in PDF format, the Pdfminer Python module was employed to extract text while simultaneously filtering out headers and footers containing administrative information based on the hospital's specific templates.

Despite these precautions, the extracted texts still contain protected health information (PHI), such as the patient's name and date of birth. Manual elimination of this information from the large volume of documents would be impractical. Consequently, the decision was made to create surrogates documents\cite{carrell2013hiding} that keep the useful information from the originals but with fictitious PHIs. DEDUCE\cite{menger2018deduce}, a rule-based tool designed for identifying PHIs in Dutch medical texts, was adapted to work with French for this purpose\footnote{available at \url{https://github.com/aenglebert/deduced}}. 

To assess the performance of our modified DEDUCE~\cite{menger2018deduce} method used with french documents, 100 reports were randomly selected in the dataset and manually annotated for patient names, person names, locations, institutions, dates, ages, id numbers, phone numbers and url/e-mails. The proposed method was then compared with the annotations, the precision, recall and F1-score were computed for each PHI with results available in Table~\ref{tab:pseudonymisation}

After identifying PHI in the documents, a systematic replacement was carried out using fictitious but contextually coherent substitute names for individuals, locations, and health institutions.
To ensure authenticity in the surrogate data, last names and first names were sourced from the Belgian \textit{Direction générale Statistique} (StatBel)\footnote{https://statbel.fgov.be/fr/themes/population/noms-et-prenoms} and the French \textit{Institut national de la statistique et des études économiques} (INSEE). For health institution names, lists of nursing homes and hospitals from the Belgian \textit{Institut national d'assurance maladie invalidité} (INAMI) were used. A list of all cities in Belgium provided addresses. To further protect privacy, a random shift (between -1000 and +1000 days) was applied to dates, while phone numbers, URLs, and email addresses were simply removed.

The creation of surrogates using the above pseudonymization process is crucial to ensure adherence to privacy regulations, notably GDPR\cite{GDPR}. Furthermore, the incapacity to categorize pseudonymized data as entirely anonymized imposes a limitation on our capacity to publicly share the dataset.

This complete process yielded a dataset of 1,837,427 surrogates of medical documents.

\subsection{Images and reports pairing}\label{sec:pairing}

To conduct contrastive vision-language pretraining, pairs of image and text data need to be made.
The documents were restricted to radiology reports and aligned with X-ray studies based on their dates (before pseudonymization). In cases where multiple studies and X-ray reports exist for a specific date, we align them in chronological order while disregarding ambiguous instances that necessitate manual examination.

The resulting number of paired studies amounts to 219,675, corresponding to 789,397 individual X-ray images in total. As a study may consist of one or several images, there exists a one-to-many relationship between the reports and the images.

 It's worth noting that only pairs of radiographs with associated reports were utilized, which represents only a fraction of the available reports presented in Section~\ref{sec:reports}. For instance, reports of orthopedic consultations were excluded since they are not directly paired to the images, even though they may contain valuable related information not present in radiologists' reports.

\begin{table}[]
    \centering
\begin{tabular}{lrlll}
\toprule
\textbf{}     & \textbf{Count} & \textbf{Precision} & \textbf{Recall} & \textbf{F1-score} \\
\midrule
Patient names & 132.0          & 0.96           & 1.00        & 0.98          \\
Person names  & 100.0          & 0.66           & 0.94        & 0.78          \\
Locations     & 52.0           & 0.98           & 0.86        & 0.92          \\
Institutions  & 23.0           & 0.76           & 0.83        & 0.79          \\
Dates         & 427.0          & 0.99           & 0.98        & 0.98          \\
Ages          & 39.0           & 0.86           & 0.97        & 0.91          \\
ID numbers    & 19.0           & 0.95           & 1.00        & 0.97          \\
Phone numbers & 47.0           & 0.98           & 0.93        & 0.96          \\
URL/e-mails   & 13.0           & 1.00           & 1.00        & 1.00         \\
\bottomrule

\end{tabular}
    \vspace{0.2cm}
    \caption {PHI identification metrics}
    \label{tab:pseudonymisation}
\end{table}

\section{Supervised validation dataset}\label{app:val_dataset}

\subsection{Dataset creation}
This dataset was constructed using images not seen by the model during VLP. The 4096 reports with their related X-Rays left behind as a validation set were reused to this purpose. 

To produce a supervision, the reports were processed using a Llama 3 70B model by prompting the model to produce a binary bone fracture classification. It was explicitly asked to the model to ignore ambiguous classifications. Additionnaly, the process was performed 3 times using a top p sampling with probability set to 0.95 with a softmax temperature of 0.8, we then keep only the labeled reports with consistent results for the three runs, resulting in 3802 labels. We then randomly sampled to keep only one labeled report per patient, resulting in 1351 labeled studies.

A test set was reserved, containing 256 labeled studies. The remaining 1095 studies are used for the validation and train set and are dynamically split during the experiments.

\subsection{Additional figure from experiments}\label{app:osrx_vlp_all}

\begin{figure}[h!]
    \centering
    \includegraphics{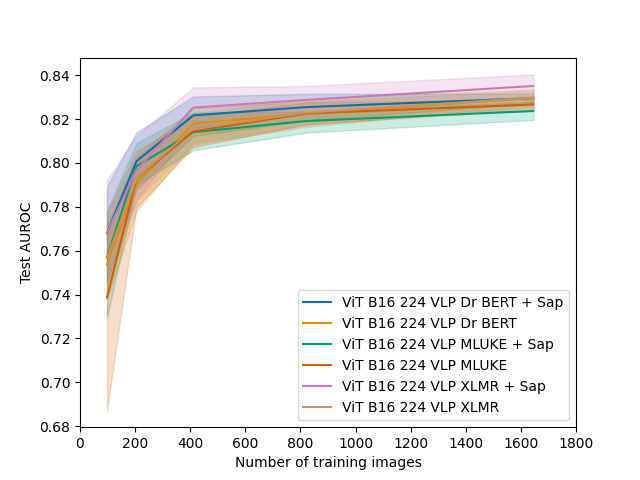}
    \caption{Classification results achieved using a \textbf{linear projection of a frozen vision encoder} on varying numbers of images obtained from the same hospital as the pretraining dataset. The shaded areas around the lines represent the 95\% confidence intervals, calculated from 8 training sessions for each specified number of training images, with different seeds used for sampling these images. This additional figure show the performances of all the VLP models. ImageNet and random models were removed for better view. See Figure~\ref{fig:osrx_plot} for comparison with ImageNet and Random models.}
    \label{app:fig:osrx_vlp_all}
\end{figure}

Figure~\ref{app:fig:osrx_vlp_all} show ta comparison between all the VLP models on the validation dataset for experiment in Section~\ref{sec:exp:supervised_validation}.

\section{Datasets for downstream tasks}\label{app:down_datasets}

To asses the performances achieved on multiple downstream tasks, we considered the following datasets (also summarized in Table~\ref{tab:datasets}):

\begin{itemize}
\item \textbf{MURA\cite{rajpurkar2017mura}} The dataset is composed of 40,561 radiographs of the upper arm from 14,863 studies, with each study labeled as normal or abnormal. The test set of 207 images is not publicly available. To facilitate comparison with previous works~\cite{Zhang2022convirt}, we mimic the same strategy by using the original validation set of 3,197 images as a test set and randomly split the original training set of 36,808 images into a validation set and a new training set.
For the retrieval task, since no comparison with previous work was available, a 5 folds has been adopted to allow computation of a standard deviation.

\item \textbf{FracAtlas\cite{abedeen2023fracatlas}} The dataset includes 4,083 images annotated for bone fracture classification, localization and segmentation. In this work, the classification annotations were used. We randomly split the dataset using 3,267 images for training, 408 images as a validation set, and 408 images for testing purpose. For the retrieval task, the whole dataset was split into 5 random folds.

\item \textbf{Osteoarthritis Initiative\cite{nevitt2006osteoarthritis} - Kellgren-Lawrence scale}:
The Osteoarthritis Initiative (OAI) is a ten-year observational study conducted across multiple centers, encompassing a total of 4,796 subjects. It comprises a diverse array of datasets, encompassing both imaging modalities such as radiographs and MRI scans, as well as clinical measurements.

One notable dataset within the study contains data scored according to the Kellgren-Lawrence (KL) semi-quantitative scale, applied to knee radiographs obtained at various intervals throughout the study period. The KL scale grades the severity of osteoarthritis, utilizing a five-point scale ranging from 0 to 4.

For the purpose of our classification task, we focused on knee radiographs obtained at the 12, 24, 36, and 48-month visits, for which KL scores are available.

A total of 16,480 radiographs, corresponding to 32,960 knees, are considered for this downstream task. We then partitioned these images into three distinct sets: a training set comprising 13,145 images, a validation set comprising 1,667 images, and a test set with 1,668 images. To maintain data independence and avoid potential biases, we ensured that images from the same patient were not distributed across different sets.

\item \textbf{RSNA Pediatric Bone Age \cite{halabi2019rsna}} This dataset features pediatric hand radiographs paired with patient age in months. The dataset encompasses a total of 14,236 hand radiographs, with 12,611 allocated to the training set, 1,425 to the original validation set, and 200 to the test set. As the test set is not publicly available, we repurposed the original validation set to serve as the test set. Additionally, a new validation set was defined using 10\% of the original training set, while the remaining 90\% are used as the final training set.

\item \textbf{Osteoarthritis Initiative\cite{nevitt2006osteoarthritis} - Hip-Knee-Ankle angle}: Within the OAI project, there also exists a dataset containing measurements of the Hip-Knee-Ankle (HKA) angle derived from full limb radiographs. Specifically, we utilized full limb radiographs obtained at the 12, 24, 36, and 48-month visits, each accompanied by corresponding HKA angle measurements.

This compilation yields a total of 3,783 images, which we divided into distinct subsets for training, validation, and testing. The training set comprises 3,027 images, while the validation and test sets consist of 381 and 375 images, respectively. As was done with the KL dataset, we ensured that patients were allocated to only one set to prevent the presence of multiple images from the same patient across different subsets.

\end{itemize}

\section{Zero-shot Strategies}\label{app:zero_strategies}

\subsection{MURA}\label{app:mura_zero}

\begin{itemize}
    \item \textbf{Text binary}: We used a simple "abnormal" vs "normal" text prompt. The exact prompts are, in French, "normal" and "anormal".
    \item \textbf{Text enumeration}: The text prompt for the abnormal label consists of a comma-separated enumeration of the pathologies considered as abnormal by the authors of the MURA dataset. The used prompts are "fracture, luxation, arthrose, ostéosynthèse, arthroplastie" and "normal".
    \item \textbf{Latent minimum}: Multiple text prompts are used as sub-classes for each pathology in the abnormal label (same pathologies as text enumeration, but each in a different prompt), each encoded in the latent space. The prompt for the pathology with the minimum cosine distance to each image embedding is used. The used prompts are "fracture", "luxation", "arthrose", "ostéosynthèse", "arthroplastie" and "normal".
    \item \textbf{Latent mean}: Similar to Latent minimum, the same sub-classes are used and encoded in the latent space. The average of all abnormal prompts embedding is used instead of the closest pathological prompt. The used prompts are "fracture", "luxation", "arthrose", "ostéosynthèse", "arthroplastie" and "normal".
\end{itemize}

\subsection{FracAtlas}\label{fracatlas_zero}

\begin{itemize}
    \item \textbf{Text binary}: We used a "fracture" vs "normal" text prompt. The exact prompts are, in French, "fracture" and "normal".
    \item \textbf{Text enumeration}: The text prompt for the bone fracture label consists of a comma-separated enumeration of bone fracture associated terms. The used prompts are "fracture, lésion osseuse, arrachement osseux" and "normal".
    \item \textbf{Latent minimum}: The terms used in the text enumeration are encoded one at a time and the  pathological prompt with the minimum cosine distance from the evaluated image embedding is used as reference for the fracture label.
    The prompts are  "fracture", "lésion osseuse", "arrachement osseux" and "normal".
    \item \textbf{Latent mean}: The terms used in the text enumeration are encoded one at a time and averaged for the fracture label.
    The prompts are  "fracture", "lésion osseuse", "arrachement osseux" and "normal".
\end{itemize}

\section{Additional Latent space Figures}\label{app:latent_additional}

This section contains the t-SNE and LDA as shown in Section~\ref{sec:latent} in Figure~\ref{fig:tsne_mura_loc} and Figure~\ref{fig:lda_mura} for all the models explored. 

\begin{figure}
    \centering
    \begin{subfigure}{0.32\linewidth}
        \includegraphics[width=\linewidth]{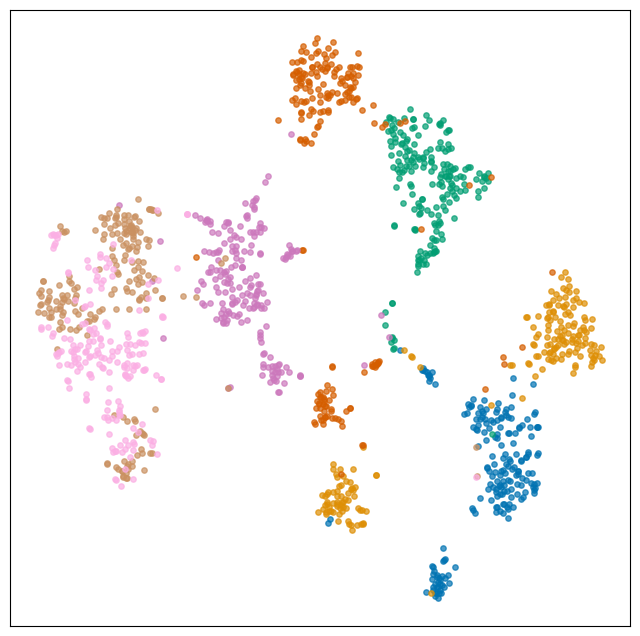}
        \caption{VLP with Dr BERT}
    \end{subfigure}
    \begin{subfigure}{0.32\linewidth}
        \includegraphics[width=\linewidth]{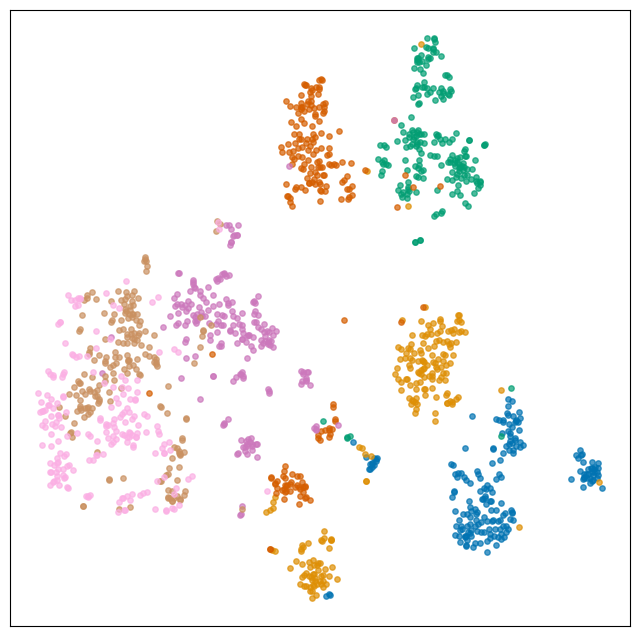}
        \caption{VLP with Dr BERT + Sap}
    \end{subfigure}
    \begin{subfigure}{0.32\linewidth}
        \includegraphics[width=\linewidth]{figures/tsne/tsne_mura_locations_checkpoints_vit_b16_224-mluke_.png}
        \caption{VLP with MLUKE}
    \end{subfigure}
    \begin{subfigure}{0.32\linewidth}
        \includegraphics[width=\linewidth]{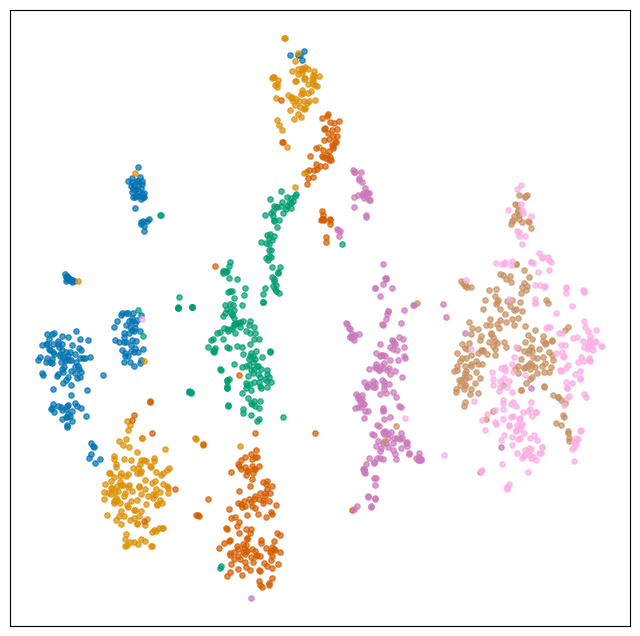}
        \caption{VLP with MLUKE + Sap}
    \end{subfigure}
    \begin{subfigure}{0.32\linewidth}
        \includegraphics[width=\linewidth]{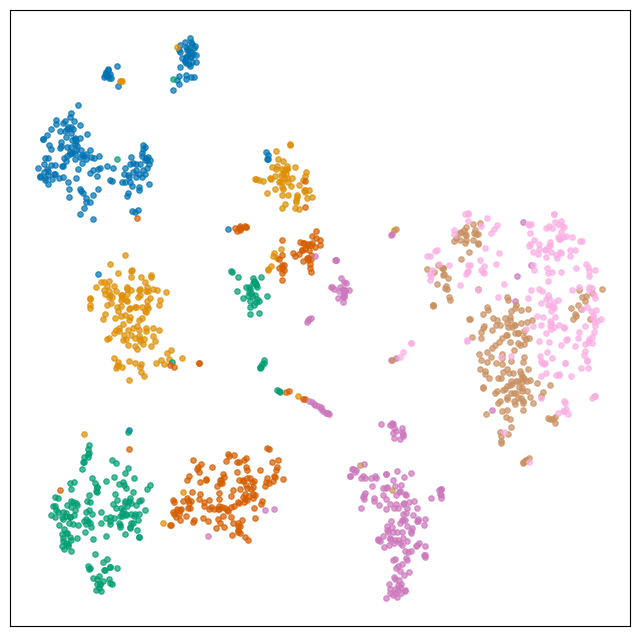}
        \caption{VLP with XLMR}
    \end{subfigure}
    \begin{subfigure}{0.32\linewidth}
        \includegraphics[width=\linewidth]{figures/tsne/tsne_mura_locations_checkpoints_vit_b16_224-sapbert_xlmr_.png}
        \caption{VLP with XLMR+Sap}
    \end{subfigure}
    \begin{subfigure}{0.32\linewidth}
        \includegraphics[width=\linewidth]{figures/tsne/tsne_mura_locations_imagenet.png}
        \caption{no VLP (ImageNet)}
    \end{subfigure}
    \caption{t-SNE visualizations of the embeddings of MURA images with and without VLP pretraining for the anatomical locations. No training on MURA was conducted for any of the models}
    \label{fig:tsne_mura_location_full}
\end{figure}

\begin{figure}
    \centering
    \begin{subfigure}{0.32\linewidth}
        \includegraphics[width=\linewidth]{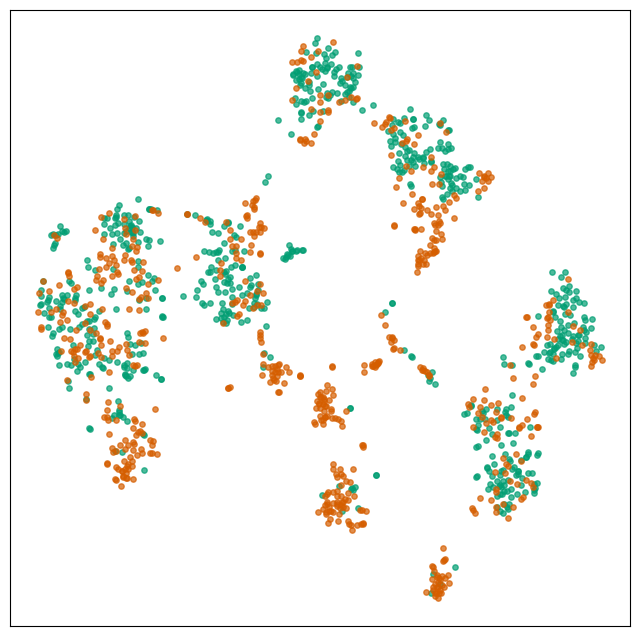}
        \caption{VLP with Dr BERT}
    \end{subfigure}
    \begin{subfigure}{0.32\linewidth}
        \includegraphics[width=\linewidth]{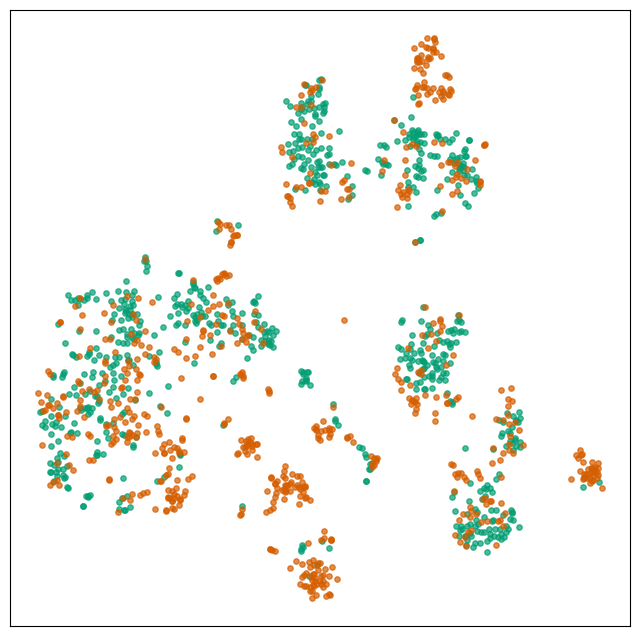}
        \caption{VLP with Dr BERT + Sap}
    \end{subfigure}
    \begin{subfigure}{0.32\linewidth}
        \includegraphics[width=\linewidth]{figures/tsne/tsne_mura_labels_checkpoints_vit_b16_224-mluke_.png}
        \caption{VLP with MLUKE}
    \end{subfigure}
    \begin{subfigure}{0.32\linewidth}
        \includegraphics[width=\linewidth]{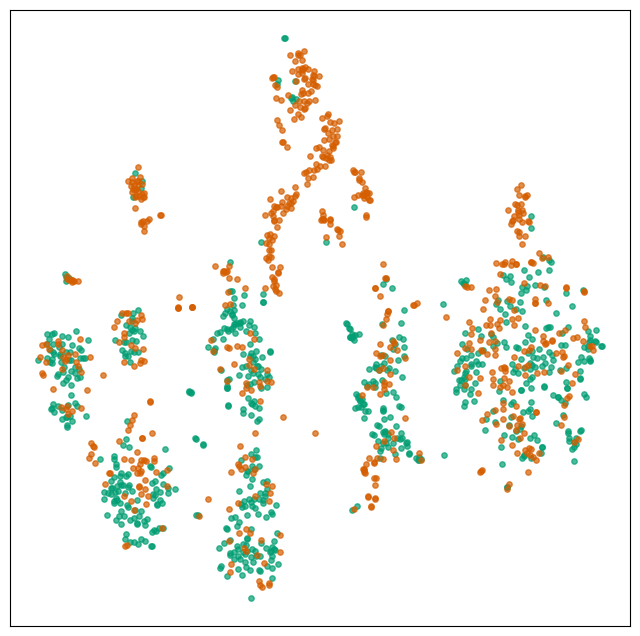}
        \caption{VLP with MLUKE + Sap}
    \end{subfigure}
    \begin{subfigure}{0.32\linewidth}
        \includegraphics[width=\linewidth]{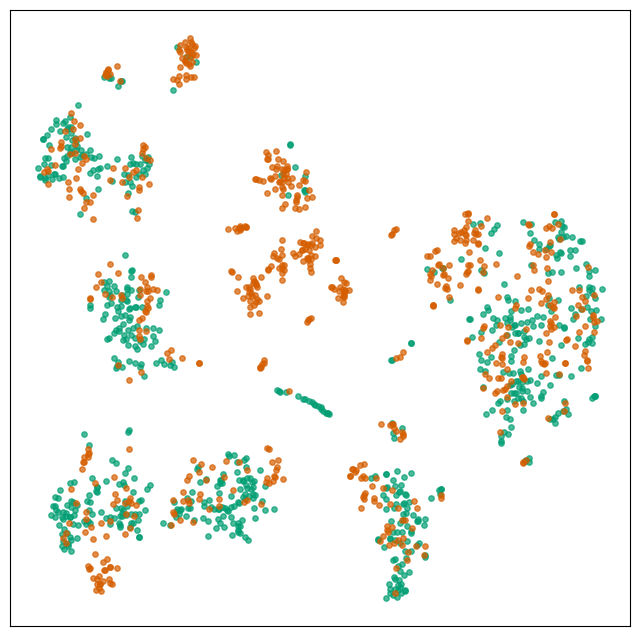}
        \caption{VLP with XLMR}
    \end{subfigure}
    \begin{subfigure}{0.32\linewidth}
        \includegraphics[width=\linewidth]{figures/tsne/tsne_mura_labels_checkpoints_vit_b16_224-sapbert_xlmr_.png}
        \caption{VLP with XLMR+Sap}
    \end{subfigure}
    \begin{subfigure}{0.32\linewidth}
        \includegraphics[width=\linewidth]{figures/tsne/tsne_mura_labels_imagenet.png}
        \caption{no VLP (ImageNet)}
    \end{subfigure}
    \caption{t-SNE visualizations of the embeddings of MURA images with and without VLP pretraining for the normal and abnormal labels. No training on MURA was conducted for any of the models.}
    \label{fig:tsne_mura_label_full}
\end{figure}

\begin{figure}
    \centering
    \begin{subfigure}{0.32\linewidth}
        \includegraphics[width=\linewidth]{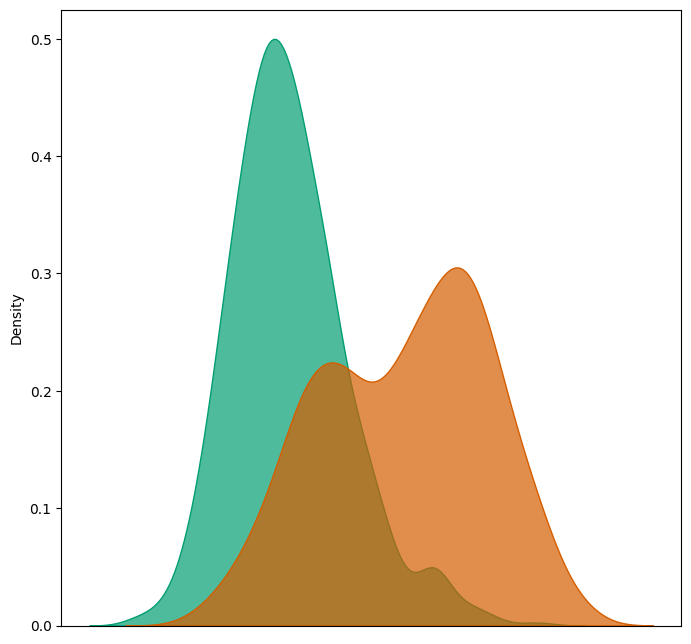}
        \caption{VLP with Dr BERT}
    \end{subfigure}
    \begin{subfigure}{0.32\linewidth}
        \includegraphics[width=\linewidth]{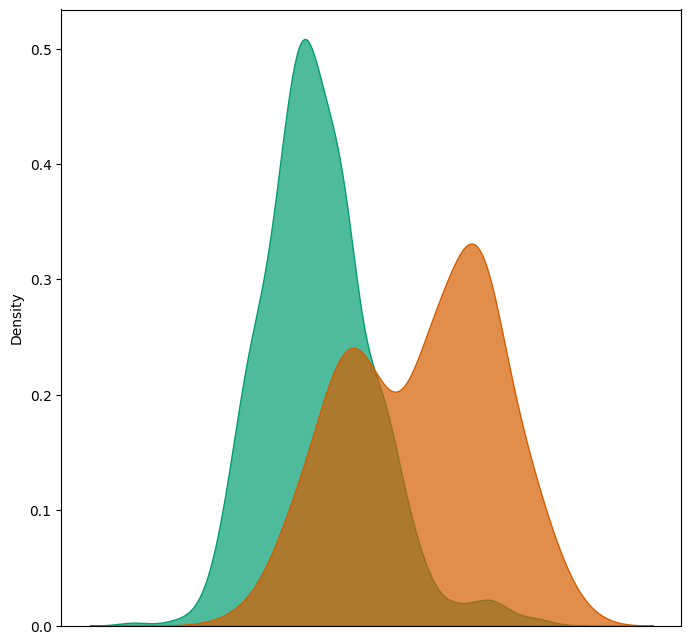}
        \caption{VLP with Dr BERT + Sap}
    \end{subfigure}
    \begin{subfigure}{0.32\linewidth}
        \includegraphics[width=\linewidth]{figures/lda/lda_mura_labels_checkpoints_vit_b16_224-mluke_.png}
        \caption{VLP with MLUKE}
    \end{subfigure}
    \begin{subfigure}{0.32\linewidth}
        \includegraphics[width=\linewidth]{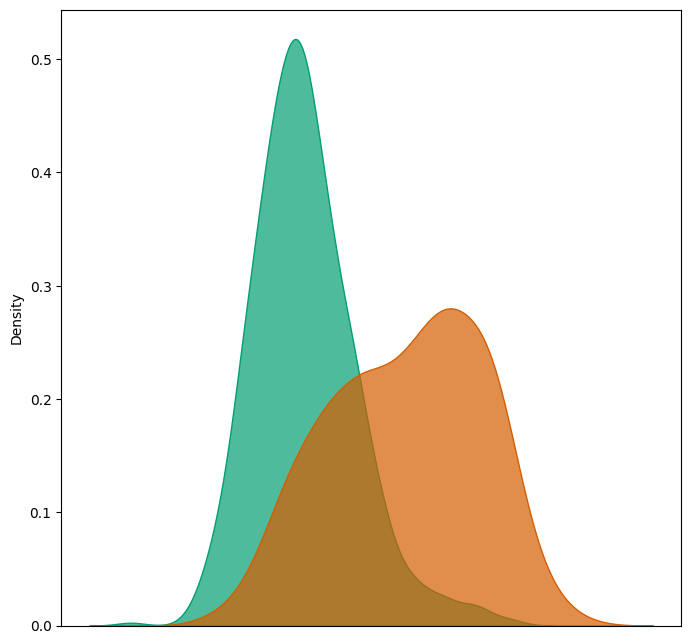}
        \caption{VLP with MLUKE + Sap}
    \end{subfigure}
    \begin{subfigure}{0.32\linewidth}
        \includegraphics[width=\linewidth]{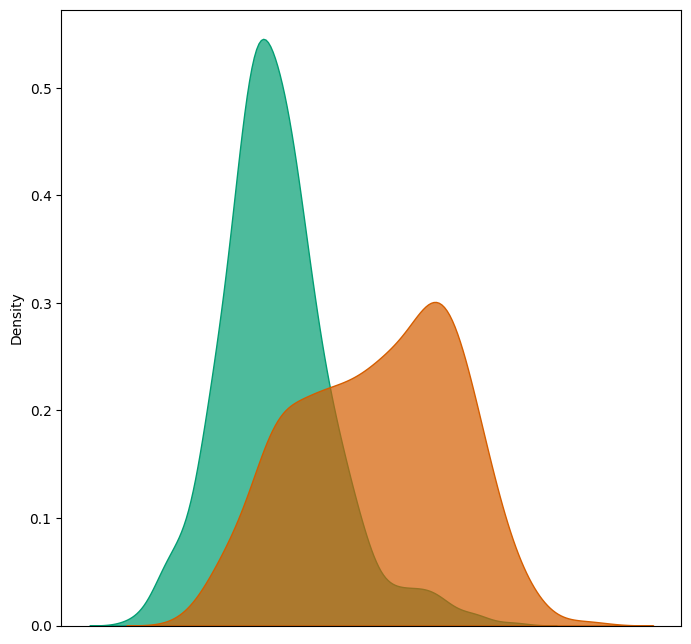}
        \caption{VLP with XLMR}
    \end{subfigure}
    \begin{subfigure}{0.32\linewidth}
        \includegraphics[width=\linewidth]{figures/lda/lda_mura_labels_checkpoints_vit_b16_224-sapbert_xlmr_.png}
        \caption{VLP with XLMR + Sap}
    \end{subfigure}
    \begin{subfigure}{0.32\linewidth}
        \includegraphics[width=\linewidth]{figures/lda/lda_mura_labels_imagenet.png}
        \caption{no VLP (ImageNet), labels}
    \end{subfigure}
    \caption{LDA visualizations of the embeddings of MURA images with and without VLP pretraining. No training on MURA was conducted for any of the models.}
    \label{fig:lda_mura}
\end{figure}


\end{document}